\documentclass[runningheads]{llncs}
\usepackage{graphicx}
\usepackage{amsmath,amssymb} 
\usepackage{subcaption}
\usepackage{xcolor}
\usepackage[width=122mm,left=12mm,paperwidth=146mm,height=193mm,top=12mm,paperheight=217mm]{geometry}
\usepackage{multirow}

\begin{document}
\pagestyle{headings}
\mainmatter
\def\ECCV18SubNumber{1816}  

\title{T-RECS: Training for Rate-Invariant Embeddings by Controlling Speed for Action Recognition} 

\titlerunning{T-RECS: Training for Rate-Invariant Embeddings by Controlling Speed}

\authorrunning{M. Ravi Ganesh, E. Hofesmann, B. Min, N. Gafoor and J. J. Corso}

\author{Madan Ravi Ganesh$^{}$ \and Eric Hofesmann$^{}$ \and Byungsu Min \and Nadha Gafoor\and \\  Jason J. Corso}
\institute{University of Michigan}
\newcommand{\acro}{T-RECS}

\maketitle

\begin{abstract}
An action should remain identifiable when modifying its speed: consider the contrast between an expert chef and a novice chef each chopping an onion.
Here, we expect the novice chef to have a relatively measured and slow approach to chopping when compared to the expert.
In general, the speed at which actions are performed, whether slower or faster than average, should \emph{not} dictate how they are recognized.
We explore the erratic behavior caused by this phenomena on state-of-the-art deep network-based methods for action recognition in terms of maximum performance and stability in recognition accuracy across a range of input video speeds.
By observing the trends in these metrics and summarizing them based on expected temporal behaviour w.r.t. variations in input video speeds, we find two distinct types of network architectures.
In this paper, we propose a preprocessing method named \acro, as a way to extend deep-network-based methods for action recognition to explicitly account for speed variability in the data.
We do so by adaptively resampling the inputs to a given model.
\acro~is agnostic to the specific deep-network model; we apply it to four state-of-the-art action recognition architectures, C3D, I3D, TSN, and ConvNet+LSTM.
On HMDB51 and UCF101, \acro-based I3D models show a peak improvement of at least $2.9\%$ in performance over the baseline while \acro-based C3D models achieve a maximum improvement in stability by $59\%$ over the baseline, on the HMDB51 dataset.
\end{abstract}

\section{Introduction}
Action recognition---classifying  the action being performed in a video clip---is a core task of video understanding that demands the proper use of temporal information to capture intrinsic video structure.
While state-of-the-art activity recognition methods show high performance across various datasets, they do not explicitly account for the speed of input videos.
It follows that speed could have an impact on the performance of these models.

\begin{figure}[t!]
\centering
\includegraphics[width=0.9\textwidth]{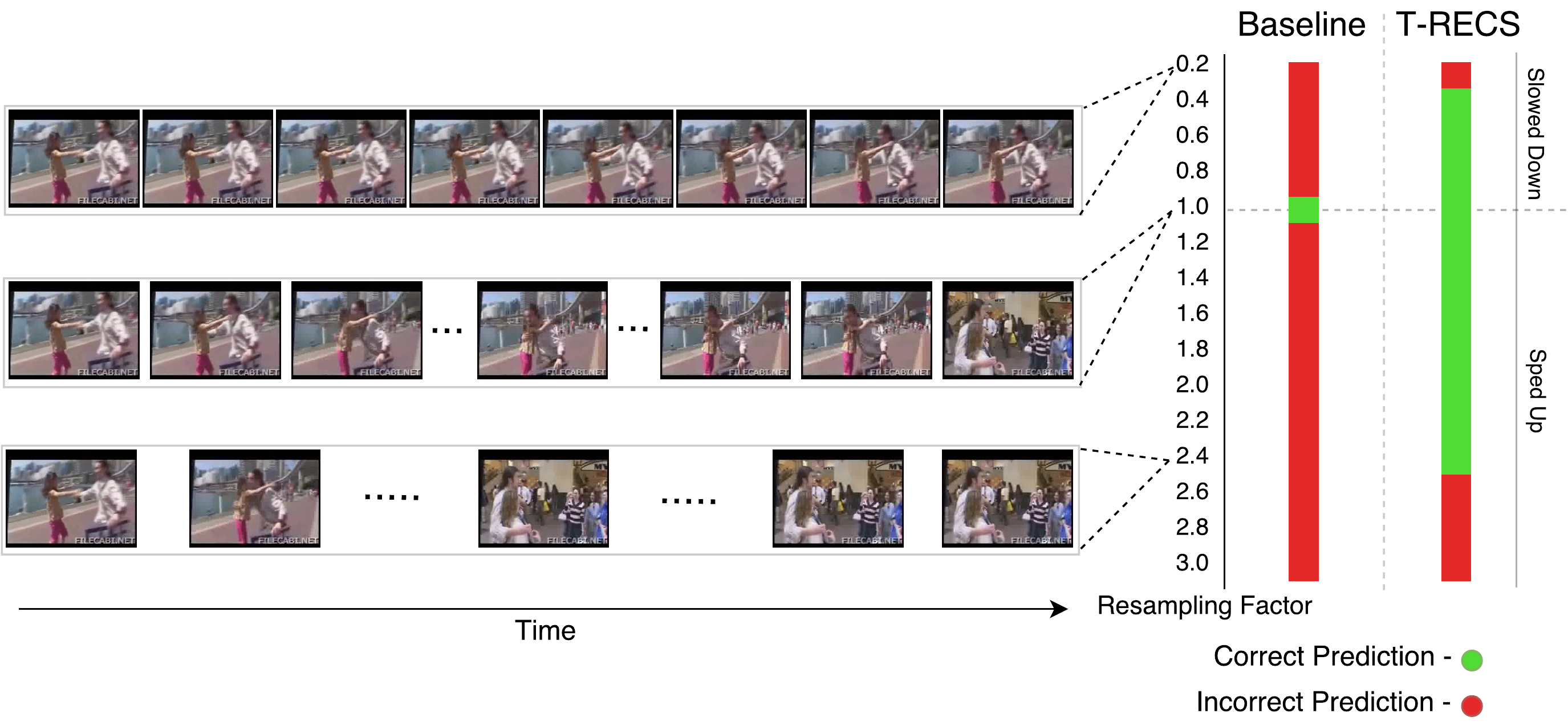}
    \caption{We show three rate modified examples of a video from the ``hug'' action class in HMDB51, each of which show a slow down, uniform sampling, and a speed up, respectively.
The graph on the side illustrates the impact of varying the speed of input videos on the recognition performance of models trained with and without \acro-based preprocessing.
The Baseline model is unable to correctly identify a sped-up or slowed-down version of the original video, which it classified correctly.
However, a variant of the baseline model trained using \acro~correctly classifies the video across a wider range of speeds.
The models used in this illustration are I3D and a \acro-modified version of I3D}
\label{fig:prob_statement}
\end{figure}

Take the example of a video from the action class ``hug,'' from the HMDB51 dataset, applied to the model I3D at varying speeds. 
From Fig.~\ref{fig:prob_statement}, we see that a simple speed up or slow down of the video is sufficient for the model to misclassify it. 
Such an example is not an anomaly; in fact, our analysis shows that misclassifications like this one occur across multiple state-of-the-art activity recognition models.
By using an empirical evaluation across these models---in which we incrementally vary the speed of input videos---we show a drop of up to $50.0\%$ in individual action class accuracy in the most extreme cases. 
This phenomena clearly points towards the need to model actions independent of the time scale at which they occur.

Classical action recognition works, such as STIP~\cite{laptev2005space}, HOG~\cite{dalal2005histograms}, and improved dense trajectories~\cite{wang2011action} do not explicitly model speed variation.
Dynamic time warping (DTW)~\cite{MyRaTASSP1981,berndt1994using} is the de facto method used in classifcal approaches to compare sequences of varying speeds.
The lack of a ground-truth canonical representation for actions renders DTW unusable for our purpose.
More recently, deep networks, including 2D two-stream approaches \cite{NIPS2014_5353}, 3D space-time approaches like Convolutional 3D (C3D)~\cite{tran2015learning}, various forms of pooling~\cite{wang2015temporal,feichtenhofer2016convolutional}, and the use of recurrent neural network (RNN) units (e.g., LSTMs, GRUs, etc.)~\cite{hochreiter1997long,cho2014properties,donahue2015long}, have shown state-of-the-art performance.
However, these deep network-based approaches along with their standard implementation practices (uniform sampling to a fixed length \cite{kay2017kinetics,donahue2015long}, streaming, and handling frames independently \cite{wang2016temporal}) tend to exploit the dataset biases as opposed to generalizing well. 

To overcome these limitations, we propose \acro, a dynamic preprocessing method during the training phase, which systematically exposes any given model to a wide range of possible speeds at which an action can occur.\footnote{The majority of motion that occurs within a frame is dominated by the subject. Hence, we use the terms ``speed of a video'' and ``speed of an action'' interchangeably.}
We do so by recycling videos from a given dataset and resampling them using an adaptive sampling rate as they feed into the network.
The adaptive resampling rate is computed based on the current cumulative number of videos used in training.
We test the generalizability to speed of input videos by testing models trained with \acro~across a range of resampling factors from $0.2$ to $3.0$ in steps of $0.2$. 
We define this testing mechanism as ``input $\alpha$ testing''.
By offering the model multiple samples of the same video at differing speeds, we encourage it to find a better minimum than the vanilla baseline, leading to higher performance.  
\acro~is not just data augmentation; \acro~enables temporally robust deep networks without the explicit need for extra data or memory.

We highlight the three main contributions put forward in this work:
\begin{enumerate}
\item We explicitly quantify the effect of speed on four state-of-the-art deep action recognition models and expose their vulnerabilities to motivate our work.
\item We design a dynamic, simple, and novel preprocessing setup that encourages models to learn \textit{more temporally robust interpretations of actions}, and we validate it on a range of speeds.
\item We define two generic categories of deep networks, based on their expected temporal behavior, by analyzing peak recognition performance and stability (which we define in Section~\ref{sec:expts}) across the baseline models and their \acro-based variants.
\end{enumerate}
Through our experiments, we find that for the action class ``jump'' in the model I3D, the maximum discrepency between the best and worst possible recognition performance, due only to a change in input video speed, is $50.0\%$ which highlights the model's vulnerability to targeted temporal attacks.
Applying \acro, we are able to reduce this to $6.7\%$ as well as achieve a peak improvement in stability of C3D by over $59\%$.
Based on our analysis of stability and peak performance, we define two broad categories with the following traits: Type I with low stability and Type II with high stability.

\section{Related works}
An accurate and effective video representation capable of capturing the intrinsic structure is important for the improvement of action recognition.
Towards this purpose, \acro~is a positive step to encourage the correct understanding and utilization of temporal information.
To the best of our knowledge we are not aware of any works that explicitly capture and illustrate the instability of current deep architectures to variations in speed of input signals nor attempt to compensate for their effects.
In this section, we highlight existing works that attempt to account for the variation in temporal scale of inputs in some form before discussing works closest in spirit to ours.

The classical approach to developing temporal understanding of data is through the use of STIP~\cite{laptev2005space}, HOG3D~\cite{klaser2008spatio}, Local Ternary Patterns~\cite{liao2010region} or SURF features generalized for videos~\cite{willems2008efficient}.
These descriptors represent snapshots of videos in a small period of time and only capture static local patterns.
They do not adequately pick up on motion features, which are critical to identifying and understanding actions.
In contrast, combining feature descriptors with tracking~\cite{wang2011action,raptis2010tracklet,lu2010learning} allows the selection of regions or objects of interest and follows their progress across a video.
Their hand-tuned nature makes them relatively less flexible to generalize across the entire spectrum of possible speeds at which an action can occur.
This combined with their inability to be learned in an end-to-end fashion makes them less suitable for the task at hand.

Turning our attention to more popular deep approaches, there are two broad categories, 1) methods that extend image-based deep architectures by incorporating temporal modeling schemes, and 2) methods that operate on the spatio-temporal domain.

Among methods that extend image-based deep architectures, there are two broad temporal modelling paradigms, 1) pooling or other aggregation procedures, and 2) the use of recurrent units.
Models that use pooling or aggregation procedures include, the original two-stream CNN~\cite{simonyan2014two} which combines predictions from independent RGB and optical flow streams, pooling features from varying spatial and temporal scales~\cite{wang2015temporal}, and one of the current state-of-the-art models that captures long-term dependencies through sparse temporal sampling~\cite{wang2016temporal}.
Although these techniques are successful in extracting and combining appearance features, they only perceive actions as a relatively static combination of appearance features repeated over certain time intervals while incoporating strong cues from semantics of the frames to boost performance levels.
On the other hand, models that use recurrent neural network units, like LSTMs and GRUs, offer an alternative approach to temporal modeling by trying to ``understand'' the sequential change in appearance features~\cite{donahue2015long}.
However, the impact of limited memory capacity and high dimensional dictionaries used to represent appearance features on the performance of recurrent units is unknown. 
This could be one of the contributing factors to their relatively weak performance when compared against other contemporary methods.
With the limited number of samples that include extreme variations in speed, RNNs fail to model temporal information to their full potential.
By exposing them to a larger number as well as a wider variety of speed variations we help them become more robust to such changes in speed.

The final modeling paradigm used in deep architectures is three dimensional convolution.
Previous works~\cite{tran2015learning,6165309} extend two-dimension neural network architectures to process spatio-temporal features through the use of three-dimensional convolutions similar to the earlier classical work by Sadanand et al.~\cite{sadanand2012action}.
Hara et. al~\cite{hara2017learning} further expand this concept to ResNet-based architectures while Carreira et al.~\cite{carreira2017quo} introduce a methodology that combines the potency of models trained on ImageNet~\cite{deng2009imagenet} with the opportunity to process spatio-temporal features in I3D.
Although such models offer the advantage of processing spatial and temporal features simultaneously, they combine features that are snapshots in time using a set of static weights.
This restricts temporal processing of data into fixed intervals and disallows the exploration of various temporal scales. 
Through our dynamic preprocessing strategy we remove the dependency that these models have on extra data to learn speed invariant characteristics by recycling data and modifying video speed on the fly.

Within existing literature, we are aware of two major works that try to tackle problems similar to our current work.
Piergiovanni~et~al.~\cite{piergiovanni2017learning} show the impact of explicitly addressing and isolating actions occurring at different temporal resolutions.
While they adapt and breakdown complex actions with no prior temporal labels, their experiments are restricted to fixed exploration schemes and do not explore robustness to varying speeds of actions.
Tallec et. al~\cite{tallec2018can} approach the task of time warping, specific to LSTMs.
Their work provides a first-principles-based solution to improving the memory capacity of LSTMs.
However, lack of ground-truth time range for actions performed in videos cobmined with the hard task of atomizing various body poses that make up an action makes the task of handling warped sequences in videos more complex.
Thus, our pursuit of an alternate strategy to address the issue of learning ``robust'' representations of actions is unique and crucial to using deep architectures to their full potential.

\section{\acro: Temporal Preprocessing}
\begin{figure}[t!]
\centering
\includegraphics[width=0.8\textwidth]{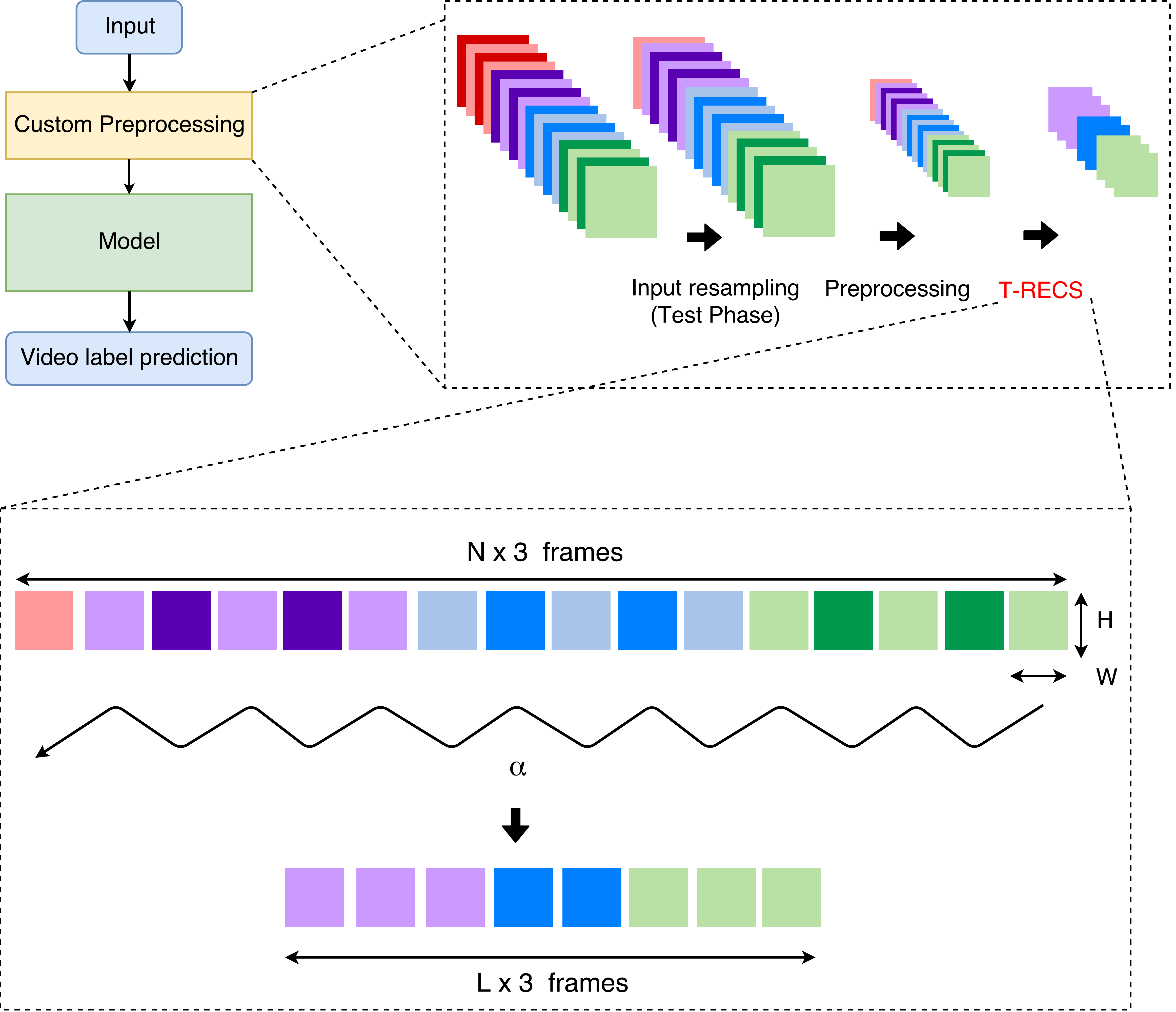}
\caption{Illustration of \acro-specific customizations to a standard input pipeline emphasizing on the simple mechanics of linear resampling using a selected $\alpha$ value (N - No. of preprocessed frames, L - Final no. of frames input to the model, H - Height of preprocessed frame, W - Width of preprocessed frame, and $\alpha$ - Resampling factor)}
\label{fig:acro}
\end{figure}

As a first step towards stabilizing overall recognition performance across varying speeds of input videos, we propose the \acro~methodology of preprocessing inputs.
We pursue networks capable of developing more robust interpretations of actions by which they maintain consistency of output predictions across temporal variations of the input.

In order to achieve this goal, \acro~takes a data-centric approach where any given model is trained by using temporally resampled input videos.
A resampling value, $\alpha$, is used to linearly resample the frames of the input video in time.
New frame indices, $l_i$, are calculated as shown in Eq.~\ref{eq:1}.

\begin{equation}
\label{eq:1}
l_i = \begin{cases} \lfloor{\alpha i}\rfloor ,& \text{if}~0 < \lfloor{\alpha i}\rfloor < N \\
                    N,& \text{if}~\lfloor{\alpha i}\rfloor = N
      \end{cases}
\end{equation}
Here, $\lfloor \rfloor$ defines the operation of rounding down a value to the nearest integer while $N$ represents the total number of input frames after preprocessing and $i \in (1,2,...,L),$ where $L$ is number of frames provided as input to the model.

By using an alternative $\alpha$ value to $1.0$, which represents uniform sampling, we allow the model to learn across multiple speeds for any given action. 
Exposing  the model to varying feature representations of an action across time scales allows us to build robustness into it.
Furthermore, the only overhead we incur is in terms of preprocessing time while avoiding excessive memory or data overages since the resampling is applied to the input video just before it is fed into the model.


Fig.~\ref{fig:acro} shows the general model training pipeline used in our experiment, with specific emphasis on the preprocessing steps. 
There are three major enhancements to our preprocessing methods when compared to common approaches,
\begin{enumerate}
\item We introduce a linear input resampling, as the first step, to help gauge the robustness of a model to varying input signals.
During the training phase, the input is unaffected by this block since its resampling factor is set to 1.0.
\item In order to avoid irregularities in information after resampling, we allow only a fixed subset of standard preprocessing functions within the customized pipeline.
They include fixed cropping, aspect preserving resize, and for models that operate on a sparse set of frames, horizontal flipping.
\item The final step of this pipeline is to resample the preprocessed input according to the adaptively chosen \acro~schedule.
\acro~$\alpha$ values are held constant during the testing phase.
\end{enumerate}

In order to explore the impact of resampling the input to a model and expose the model to extreme variations in data, we selected two incremental linear resampling strategies.
\begin{itemize}
\item \textbf{Static}: To determine the impact of resampling factors alternate to uniform sampling, we provide baselines in which the resampling factor $\alpha$ is held constant, to a value other than $1.0$, throughout the training process.
The values assigned to $\alpha$, during both training and testing, in these baselines are $0.4, 0.8, 1.5, \text{and}~2.5$.
Experiments using this resampling mechanism for preprocessing are marked by the acronym ``CVR-\textless $\alpha$\textgreater.''
\item \textbf{Dynamic}: Extending static resampling methods, we pursue two dynamic alternatives. The first is random resampling (RR) where $\alpha$ is varied randomly for every input video, where $\alpha \in \mathcal{U}(0.2, 3.0)$. The second is sinusoidal resampling (SR) where we systematically cycle through $\alpha$ values based on,
\begin{equation}
\alpha = 0.2 + 2.8\sin(v_n),
\end{equation}
where $v_n$ is the current accumulated number of videos in training.
In both cases, during testing the $\alpha$ values are set to $1.0$.
Experiments using these methods are marked by the acronym ``RR'' and ``SR,'' respectively.
\end{itemize}
Note: An artificial upper limit is imposed on these resampling methods as a safety measure to avoid skipping the critical action content in the video. 
The exact values are empirically chosen after viewing examples of resampled videos.
\section{Experiments}
\label{sec:expts}

In the following section, we provide a brief description of the datasets and baselines used within our experiments followed by an experimental subsection which details our motivating factor, results and observations, and analysis.

\subsection{Datasets}
\subsubsection{HMDB51}
HMDB51~\cite{kuehne2013hmdb51} is a collection of videos representing 51 actions.
These videos are curated from various sources including movies, the Prelinger archive, YouTube, and Google Videos.
The wide variety of sources coupled with over 101 clips for each of the 51 actions allows the HMDB51 dataset to focus extensively on intra-class variation.

\subsubsection{UCF101}
With 13,320 videos spanning 101 actions, UCF101~\cite{soomro2012ucf101} encompasses a wide variety of actions taken from videos in the ``wild.''
They include uncontrolled variations in camera motion, object appearance, illumination changes, and other real world effects. 
We do not use any of the metadata provided in the datasets.
Instead, we focus mainly on the action class videos and labels.

\subsubsection{Rate-Modified Datasets}
We resample the HMDB51 and UCF101 datasets to provide two distinct rate-modified datasets for our analysis.
In order to generate the rate-modified datasets, each video from the original dataset is resampled using ten different sampling rates.
Among these ten new videos, five are resampled using rates from~$[0.2,1.0]$ and the remaining from~$[1.0, 3.0]$.
We assume that the default speed of the action in the original video represents the unit rate, so we resample the video at varying speeds to generate the rate-modified dataset.
We use linear interpolation to generate the content of non-integer frames.

Videos generated from sampling rates $[0.2,1.0]$ represent slowing down the action being performed while those generated from sampling rates $[1.0,3.0]$ represent speeding up the action.
The training, validation, and test splits contain the same videos as the original dataset splits with the inclusion of all ten rate-modified versions generated.
The rate modified datasets are denoted as ``\textit{\textless Dataset name\textgreater Rate}'' henceforth.
\subsection{Baselines}
\subsubsection{C3D}
C3D~\cite{tran2015learning} represents an alternative modeling approach for the temporal structure of action videos by extending 2D convolutional filters to include the temporal dimension.
It does not require an explicit temporal support structure due to its ability to processes spatio-temporal features as a single unit.
However, 3D convolutional layers do not afford the benefit of ImageNet-pretraining that is granted to standard ConvNet architectures.

\subsubsection{I3D}
I3D~\cite{carreira2017quo} extends any 2D convolutional architecture to the temporal dimension by taking $N \times N$ spatial filters and repeating them over the temporal dimension $N$ times.
Thus, it jump starts its learning process using ImageNet pretrained features and offers the advantage of working with spatio-temporal representations.
The backbone of I3D used in our experiments is Inception-V1~\cite{ioffe2015batch}

\subsubsection{TSN}
Unlike most of the previous approaches, TSN~\cite{wang2016temporal} exploits sparse temporal sampling to model long-term temporal structure in an end-to-end manner.
It divides each video into K (3 in the original paper) segments and then randomly samples a snippet from each segment.
Then, a segmental consensus function is applied to the sequence of snippets to predict a video-level score.

\subsubsection{ConvNet+LSTM}
One of the most commonly used approaches for classifying action videos is by extracting frame-level features from ImageNet-pretrained standard CNN architectures, ResNet-50~\cite{donahue2015long} in our case, and applying them to a recurrent unit.
In our experiments, we use an LSTM followed by a fully connected layer. They function as the temporal modeling paradigm over frame-level features.

We provide more details with regard to the preprocessing setup used for each of the models in the Supplementary Material.
\subsection{Empirical evaluation of temporal robustness}
\begin{figure}[b!]
\centering
\includegraphics[width=\textwidth]{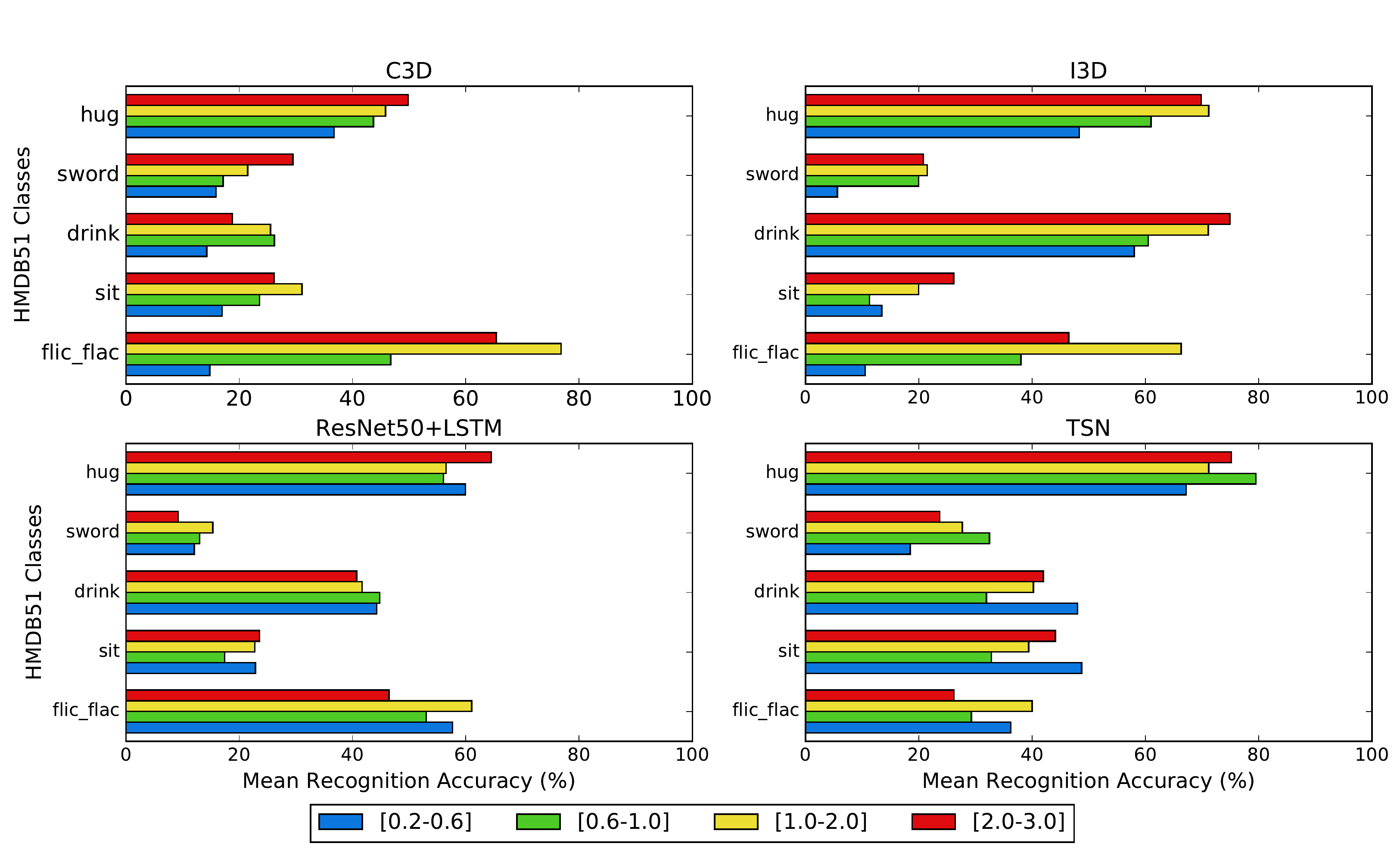}
\caption{The performance across five randomly chosen action classes in the HMDB51Rate dataset for each of the four baseline models is shown. Each graph details the variation in the model's performance as input videos are resampled using values ranging from $0.2 \text{ to } 3.0$, broken down into four bins, $[0.2, 0.6], [0.6, 1.0], [1.0, 2.0], \text{ and } [2.0, 3.0]$. Across models, the impact of variable resampling of input is unpredictable. Consider the action classes ``sword'' and ``sit'' in TSN where slowing down shows the worst and best performance respectively. These models learn a weak understanding of actions and hence do not generalize well with variable resampling (speed)}
\label{fig:action_class}
\end{figure}

We show current activity recognition models work best when actions occur at a specific speed. 
By the method that datasets are curated, there exists no fixed speed at which actions are performed. 
This inherent variation in speed of actions poses a genuine threat to the performance of the model and its applicability to videos ``in the wild.''
If an action is performed faster or slower than the model is tuned to, it can fail to classify it accurately.
In order to quantify the impact of the aforementioned problem, Fig.~\ref{fig:action_class} presents the performance of the four state-of-the-art models in study on the HMDB51Rate dataset.

From Fig.~\ref{fig:action_class}, we observe when that given the same training data, each model has a different understanding of the same action due to their architectural differences.
We clearly see that for a chosen model, there does not exist a single ``one size fits all'' speed at which it performs best on all action classes.
Furthermore, a chosen speed that increases performance in one action class might decrease performance in another.
For example, consider the action classes ``hug'' and ``sword'' in the ResNet50+LSTM model (Fig.~\ref{fig:action_class}).
We see that a speed up allows the network to achieve peak recognition perfrormance for the ``hug'' action class while having the opposite effect on the ``sword'' action class.
Further, when comparing the change in performance across different actions and models, there is no specific pattern. 
Thus, there is no one size fits all solution to the selection of speed for any model.
These problems allude to the necessity of a system that allows models to identify actions independent of the speed at which they are performed.

\subsection{\acro~Results}
Table~\ref{tab:main_expt} shows the results of applying our systematic resampling methods to our baseline models.
The accuracy values reported for each model are different to the original author provided values due to modifications in the preprocessing step to help standardize the pipeline across our models.
Accuracies were compiled for each baseline variant where the input $\alpha$ was varied from $0.2 \text{ to } 3.0$ in steps of $0.2$ (15 steps).
We refer to this as ``input $\alpha$ testing'' henceforth.
In Table~\ref{tab:main_expt}, Acc. represents the recognition accuracy result for unaltered uniform sampling of input with $\alpha = 1.0$, and Std. represents the standard deviation in performance across all fifteen resampling factors. We use Std. as a measure of the stability in performance of models.
\begin{table}[tp!]
\begin{center}
\caption{Mean recognition accuracies are shown for various SOTA action recognition models across split 1 of HMDB51 and UCF101. Accuracies are reported for each baseline model and preprocessing technique combination according to tests performed across fifteen input resampling factors ranging from $0.2 \text{ to } 3.0$ in steps of $0.2$.
Acc. represents performance on unaltered uniform sampling of input with $\alpha = 1.0$, and Std. represents the standard deviation in performance across all fifteen resampling factors. Models trained using \acro-based preprocessing almost consistently outperform their respective baselines. In general, dynamic \acro-based models show the highest stability. Results for TSN and ResNet50 + LSTM model variants show extremely high stability with HMDB51 on ResNet50 + LSTM and UCF101 on TSN show strong improvement while minimally impacting over UCF101 and HMDB51, respectively. In general, \acro-based outperform all model baselines\\}
\label{tab:main_expt}
\begin{tabular}{c|c|c|c|c|c||c|c}
\hline
\multirow{2}{*}{Base Model} & \multirow{2}{*}{Pretraining} & \multirow{2}{*}{\acro} & \multirow{2}{*}{Variant} & \multicolumn{2}{c||}{HMDB51} & \multicolumn{2}{c}{UCF101} \\\cline{5-8}
& & & & Acc (\%) & Std (\%) & Acc (\%) & Std (\%) \\
\hline
\multirow{7}{*}{C3D} & \multirow{7}{*}{Sports-1M} & -- & Baseline & 49.41 & 3.11 & 78.35 & 2.67 \\\cline{3-8}
& & \multirow{4}{*}{Static} & CVR-$0.4$ & 51.76 & 3.36 & 79.54 & 2.54 \\
& & & CVR-$0.8$ & \textbf{51.90} & 3.68 & \textbf{80.81} & 3.08 \\
& & & CVR-$1.5$ & 49.54 & 4.34 & 79.17 & 3.61 \\
& & & CVR-$2.5$ & 48.10 & 3.21 & 78.40 & 2.46 \\\cline{3-8}
& & \multirow{2}{*}{Dynamic} & RR & 49.74 & 1.45 & 79.20 & \textbf{1.18} \\
& & & SR & 49.54 & \textbf{1.26} & 79.83 & 1.34 \\
\hline
\hline
\multirow{7}{*}{I3D} & \multirow{7}{*}{Kinetics} & -- & Baseline & 60.92 & 3.55 & 88.95 & 6.00 \\\cline{3-8}
& & \multirow{4}{*}{Static} & CVR-$0.4$ & 61.63 & 8.65 & 84.69 & 9.29 \\
& & & CVR-$0.8$ & 58.82 & 5.47 & 88.13 & 6.89 \\
& & & CVR-$1.5$ & 60.26 & 2.77 & 90.22 & 3.29 \\
& & & CVR-$2.5$ & 59.02 & 3.25 & 88.82 & 2.65 \\\cline{3-8}
& & \multirow{2}{*}{Dynamic} & RR & 64.18 & \textbf{1.84} & 91.09 & 2.21 \\
& & & SR & \textbf{65.42} & 2.32 & \textbf{91.83} & \textbf{1.92} \\
\hline
\hline
\multirow{7}{*}{TSN} & \multirow{7}{*}{ImageNet} & -- & Baseline & 44.12 & \textbf{0.79} & 77.40 & 1.02 \\\cline{3-8}
& & \multirow{4}{*}{Static} & CVR-$0.4$ & 43.86 & 0.83 & 80.02 & 1.08 \\
& & & CVR-$0.8$ & \textbf{45.82} & 1.24 & 77.03 & 0.97 \\
& & & CVR-$1.5$ & 42.81 & 1.20 & 79.49 & 1.25 \\
& & & CVR-$2.5$ & 45.10 & 1.52 & \textbf{80.33} & 1.22 \\\cline{3-8}
& & \multirow{2}{*}{Dynamic} & RR & 44.38 & 1.47 & 78.14 & 1.05 \\
& & & SR & 43.79 & 0.94 & 79.22 & \textbf{0.96} \\
\hline
\hline
\multirow{7}{*}{ConvNet+LSTM} & \multirow{7}{*}{ImageNet} & -- & Baseline & 45.36 & 0.64 & 78.91 & 0.62 \\\cline{3-8}
& & \multirow{4}{*}{Static} & CVR-$0.4$ & 45.62 & 1.41 & 75.23 & 1.00 \\
& & & CVR-$0.8$ & \textbf{47.84} & 0.67 & 79.09 & 0.87 \\
& & & CVR-$1.5$ & 46.73 & 1.40 & \textbf{80.15} & 0.86 \\
& & & CVR-$2.5$ & 43.92 & 1.45 & 77.69 & 1.12 \\\cline{3-8}
& & \multirow{2}{*}{Dynamic} & RR & 47.39 & \textbf{0.61} & 76.82 & \textbf{0.45} \\
& & & SR & 47.19 & 0.68 & 78.48 & 0.59 \\
\hline
\end{tabular}
\end{center}
\end{table}

C3D generally benefits from the application of a resampling strategy during preprocessing; five of our six resampling strategies improve performance over our baselines for HMDB51 and all of our resampling strategies improve performance for UCF101.
The best performing preprocessing strategy is CVR-$0.8$ which performs at $51.90\%$ on HMDB51, $1.49\%$ above our baseline accuracy, and $80\%$ on UCF101, $2.46\%$ above our baseline accuracy.
However, the largest boost in performance through \acro-based training, in both datasets, is achieved for I3D. 
The best performing I3D variant, SR, improves by $4.5\%$ over the baseline at $65.42\%$ on HMDB51.
Both preprocessing strategies, which use dynamic $\alpha$ values, improve performance over the baseline by at least $2.14\%$ across both datasets.

TSN offers an alternate narrative for performance with an extremely high level of stability.
The maximum increase in performance of all TSN models on HMDB51 is $1.5\%$ whereas the increase in performance of most TSN models on UCF101 is over $1.5\%$.
We explore these characteristics further in the following section.

ConvNet+LSTM shows a similarly high level of stability with the worst Std. in performance being $1.45\%$ across both datasets, which is significantly lower than in variants of C3D or I3D.
The main point of contention between TSN and ConvNet+LSTM models is the similarity in performance and stability between TSN models on HMDB51 and ConvNet+LSTM models on UCF101 and the similarity between TSN model on UCF101 and ConvNet+LSTM models on HMDB51.

To summarize the results, the common thread that connects all these models is the fact that in most cases, our resampling strategies perform much better than the baseline.
Considering that stability is the preferable trait while keeping performance to acceptable levels, SR and RR strategies yield the lowest standard deviation with performances higher than the baseline in the majority of models and datasets.
This in turn points to their effectiveness in retaining performance while stabilizing the model from the influences of speed variations.
\begin{figure}[t!]
  \centering
  \begin{subfigure}[h]{0.49\textwidth}
    \centering
    \includegraphics[width=\textwidth]{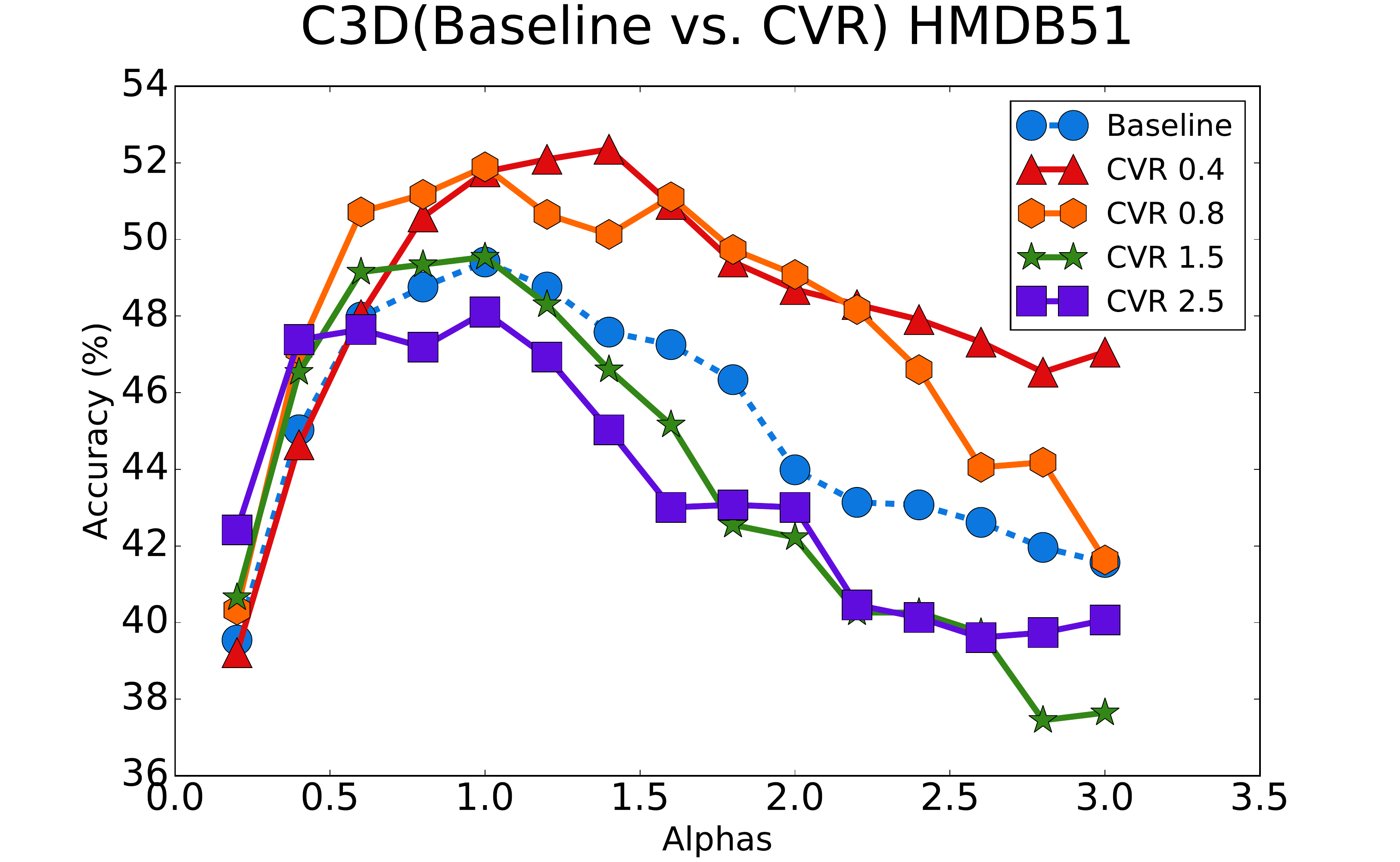}
  \end{subfigure}
  \begin{subfigure}[h]{0.49\textwidth}
    \centering
  \includegraphics[width=\textwidth]{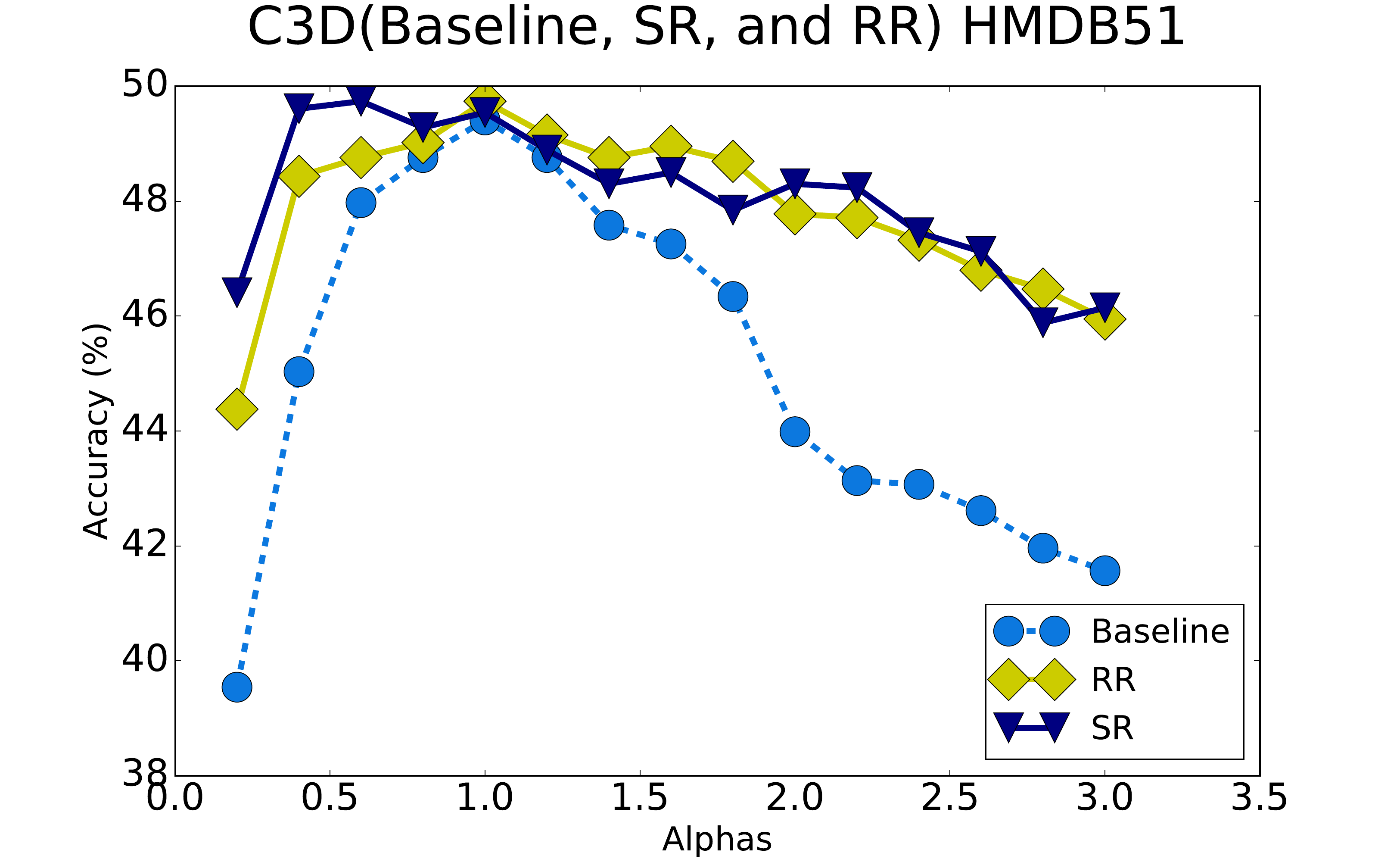}
  \end{subfigure}
  \begin{subfigure}[h]{0.49\textwidth}
    \centering
    \includegraphics[width=\textwidth]{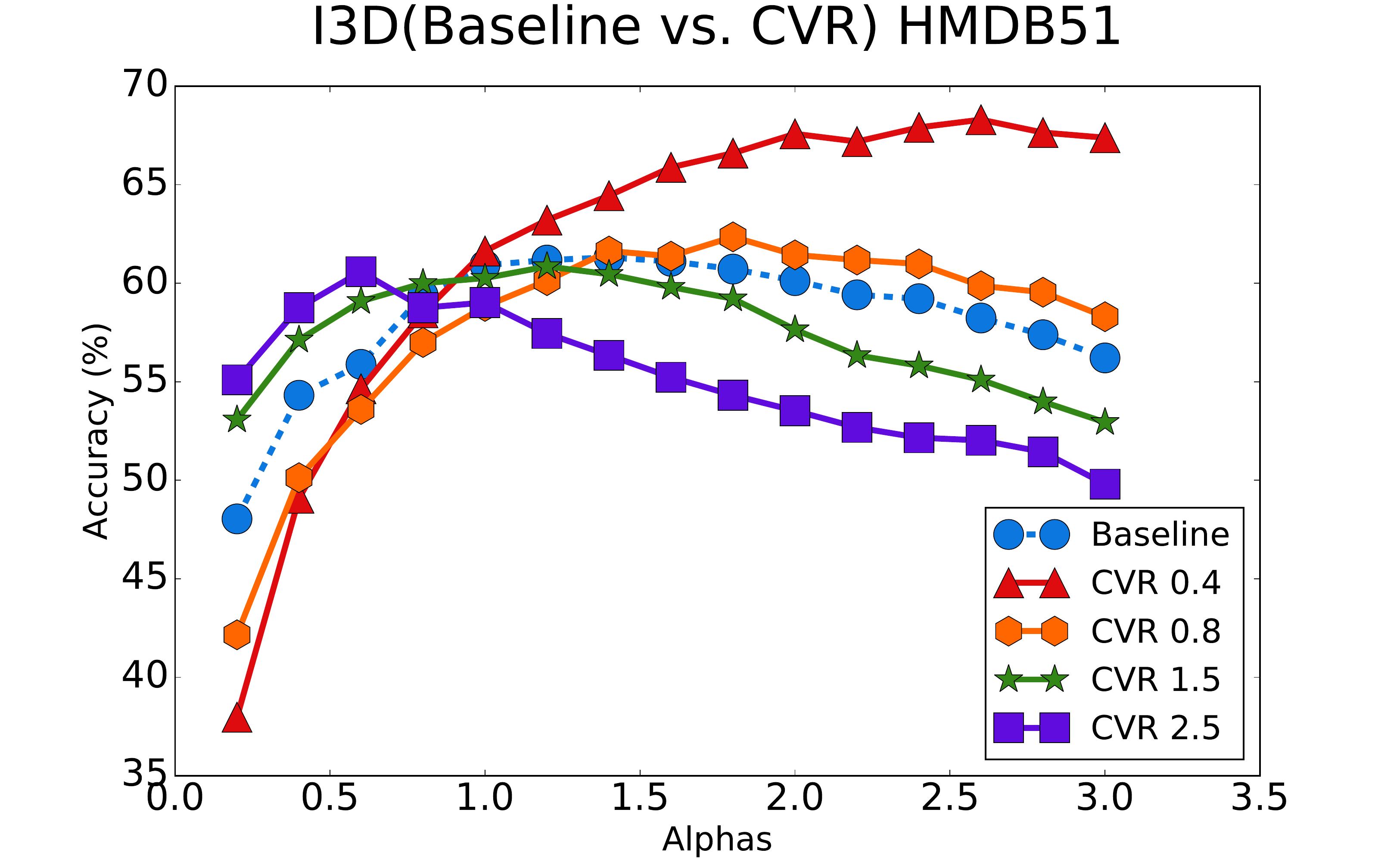}
   \end{subfigure}
  \begin{subfigure}[h]{0.49\textwidth}
    \centering
    \includegraphics[width=\textwidth]{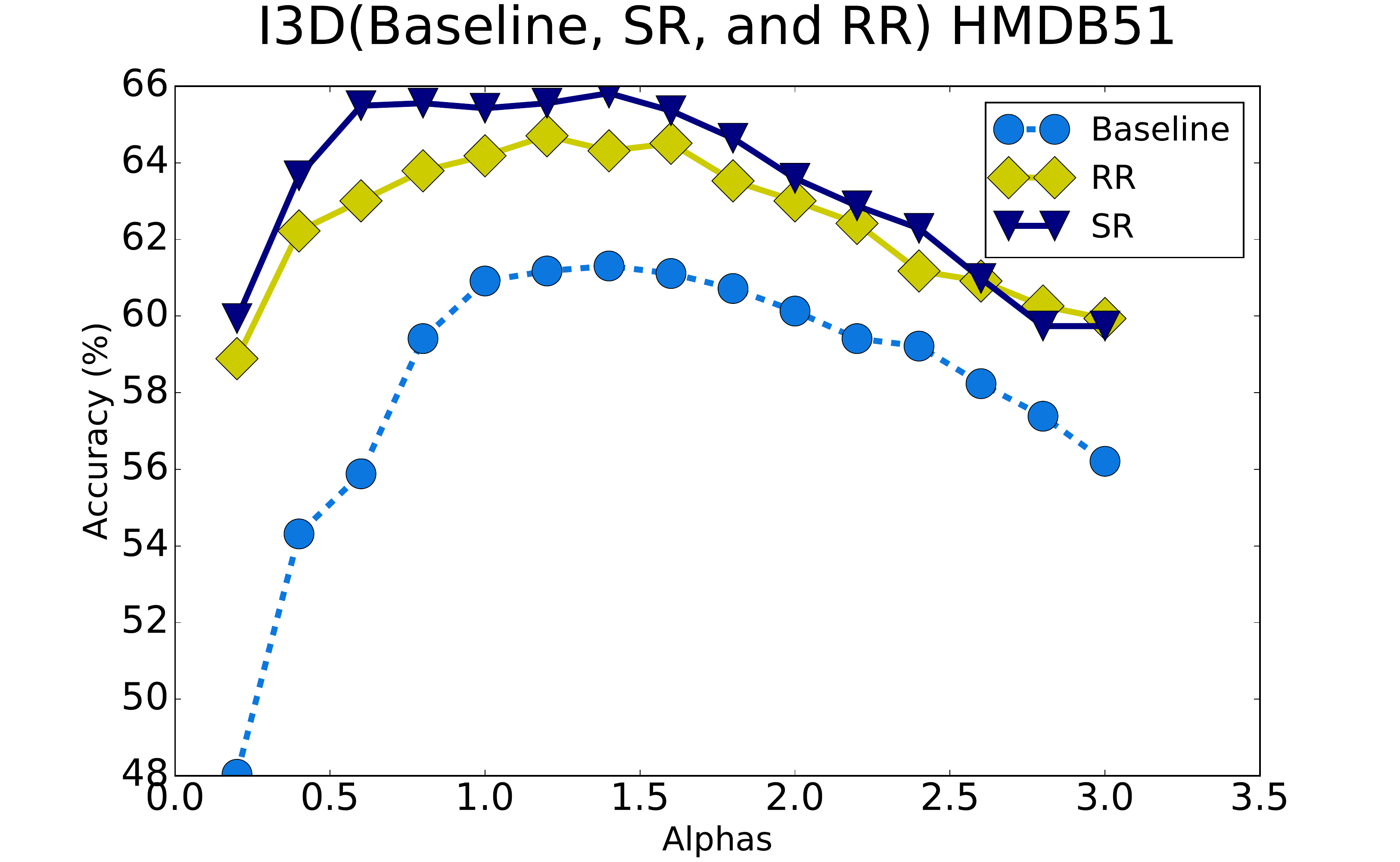}
   \end{subfigure}
  \caption{Stability in performance across multiple input resampling factors for C3D and I3D are shown. Both models are characterized by a peaked performance distribution which is retained in static \acro-based variants. However, the stability is significantly improved in dynamic \acro-based variants. These are typically expected characteristics of Type I models}
  \label{fig:c3d_i3d_alpha_testing}
\end{figure}

\section{Characterizing temporal behaviors of models}
Previous results have shown the ability of our models to improve performance over the original dataset by leveraging artificially generated variations in speed of input data.
In order to explore and characterize the expected temporal behavior of models with and without \acro-based preprocessing, we draw a direct comparison in performance using the detailed results obtained from input $\alpha$ testing.
The top row of Fig.~\ref{fig:c3d_i3d_alpha_testing} shows C3D results for input $\alpha$ testing. In general, CVR-0.4 and 0.8 have the highest performance over the majority of $\alpha$ values.
However, when we observe the extreme ends of the range of tested input $\alpha$ values, an interesting pattern emerges.
For an input $\alpha = 0.2$, CVR-1.5 and 2.5 perform better than CVR-0.4 and 0.8.
We believe that at input $\alpha = 0.2$, the video is severly undersampled and this effect is further exacerbated by the CVR-0.4 and 0.8 models.
However, for the high CVR models, these slowed-down videos are then sped up by a factor of $1.5$ or $2.5$ leading to a final speed that is closer to the standard input speed of $1.0$.
The same effect occurs at the other end of the spectrum, $\alpha = 3.0$, where CVR-0.4 and 0.8 models perform best. 
I3D shows a more exaggerated example of the same rate-cancelling effect.

\begin{figure}[t]
  \begin{subfigure}[h]{0.49\textwidth}
    \centering
    \includegraphics[width=\textwidth]{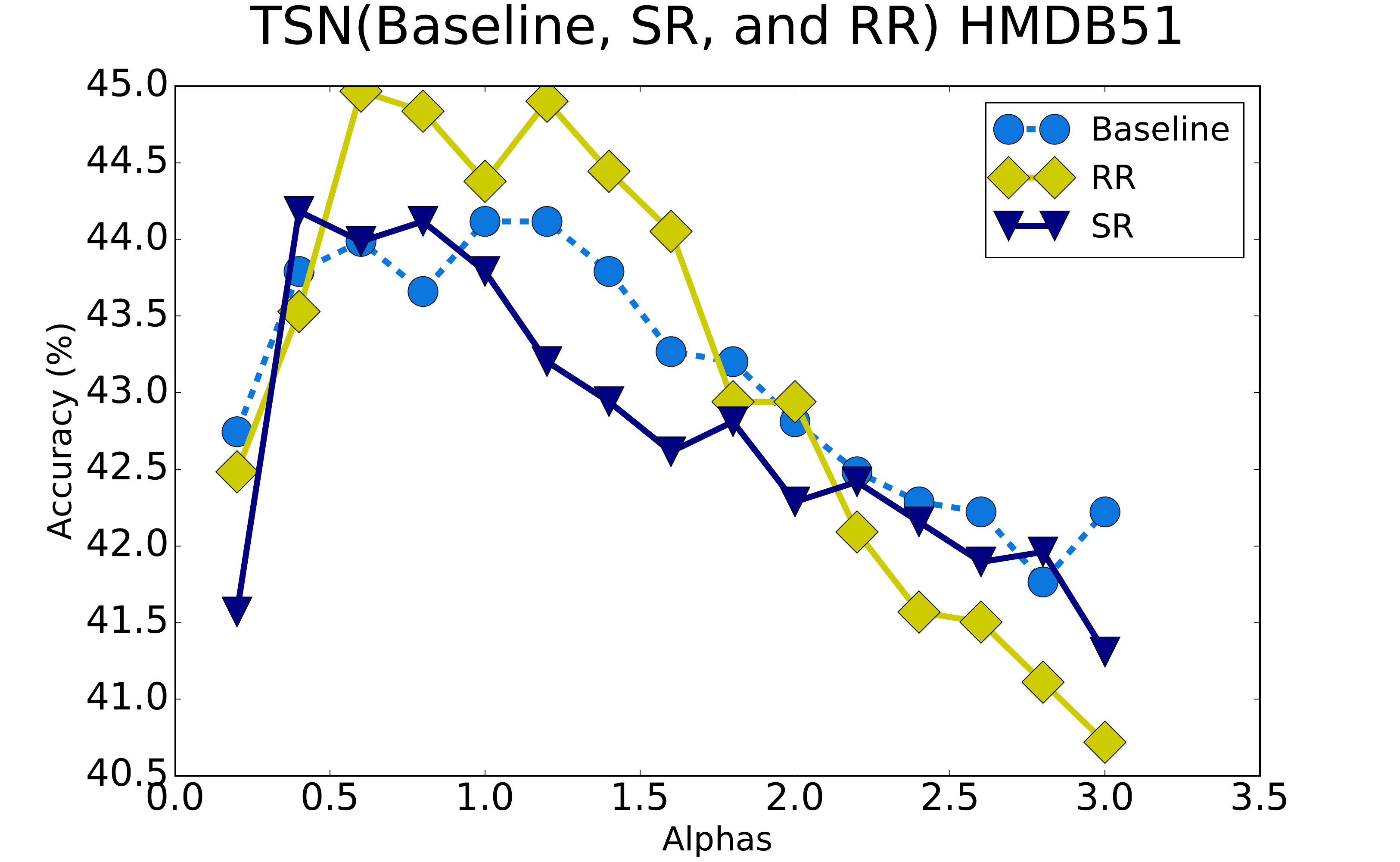}
  \end{subfigure}
  \begin{subfigure}[h]{0.49\textwidth}
    \centering
    \includegraphics[width=\textwidth]{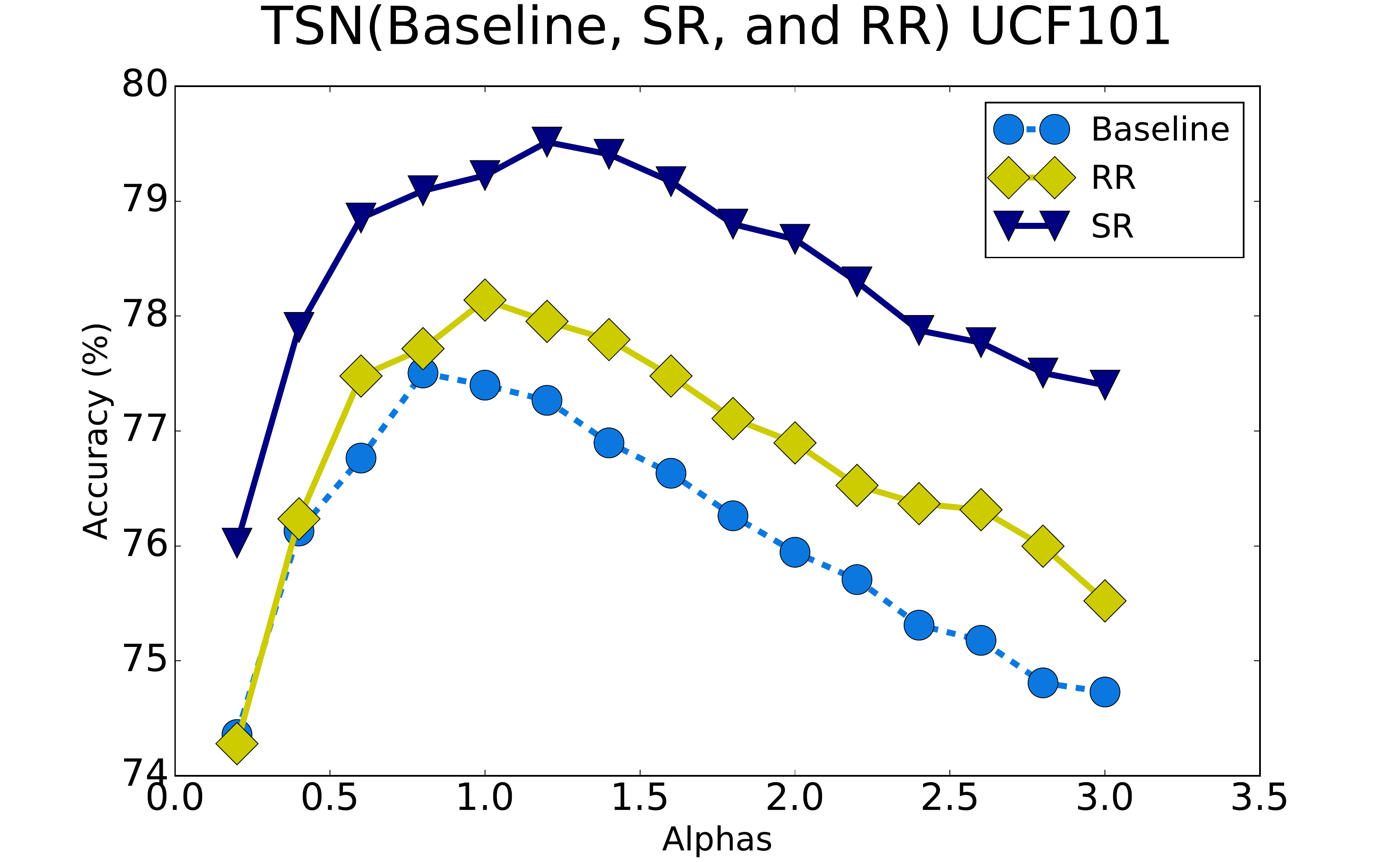}
  \end{subfigure}
  \begin{subfigure}[h]{0.49\textwidth}
    \centering
    \includegraphics[width=\textwidth]{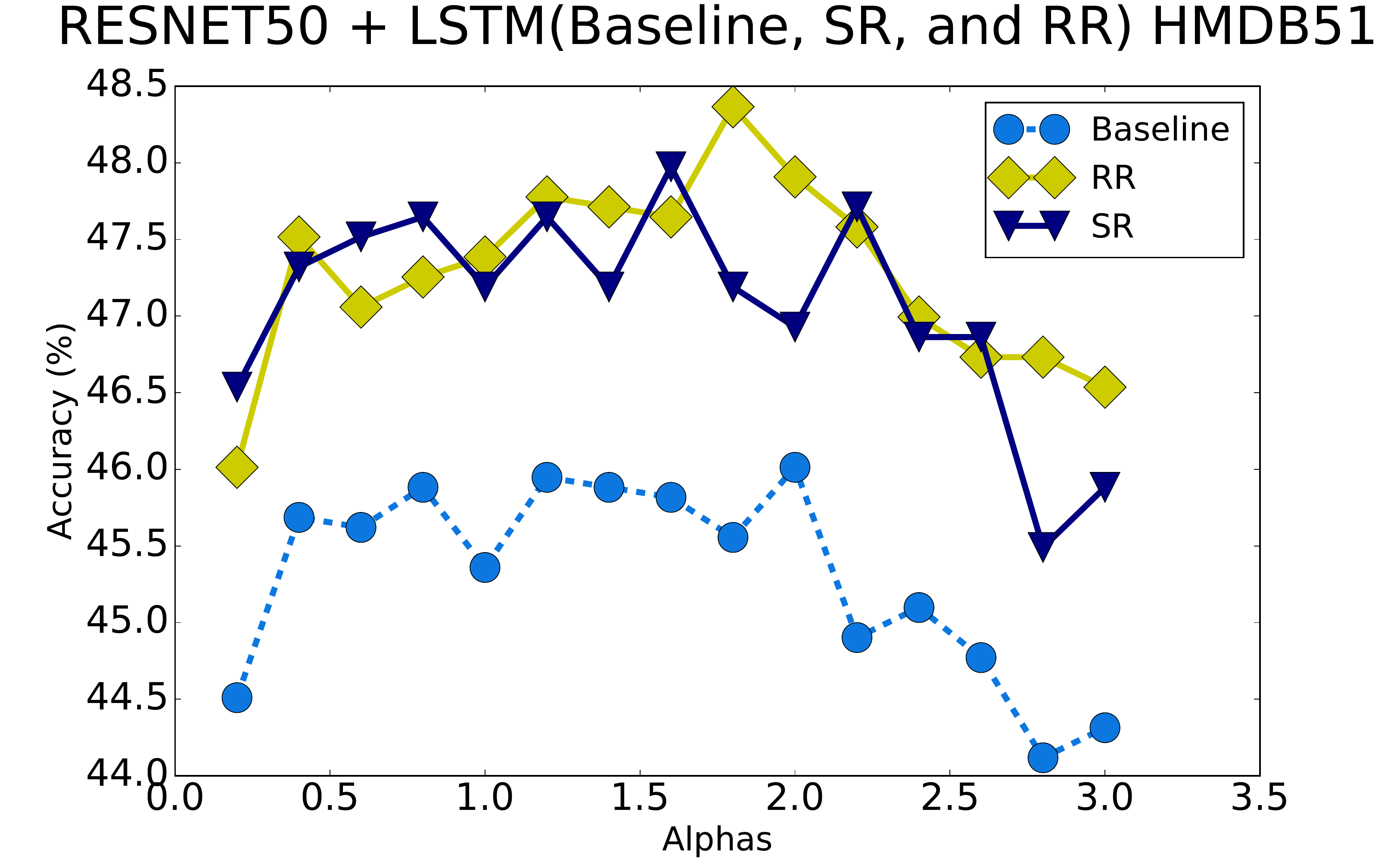}
  \end{subfigure}
  \begin{subfigure}[h]{0.49\textwidth}
    \centering
    \includegraphics[width=\textwidth]{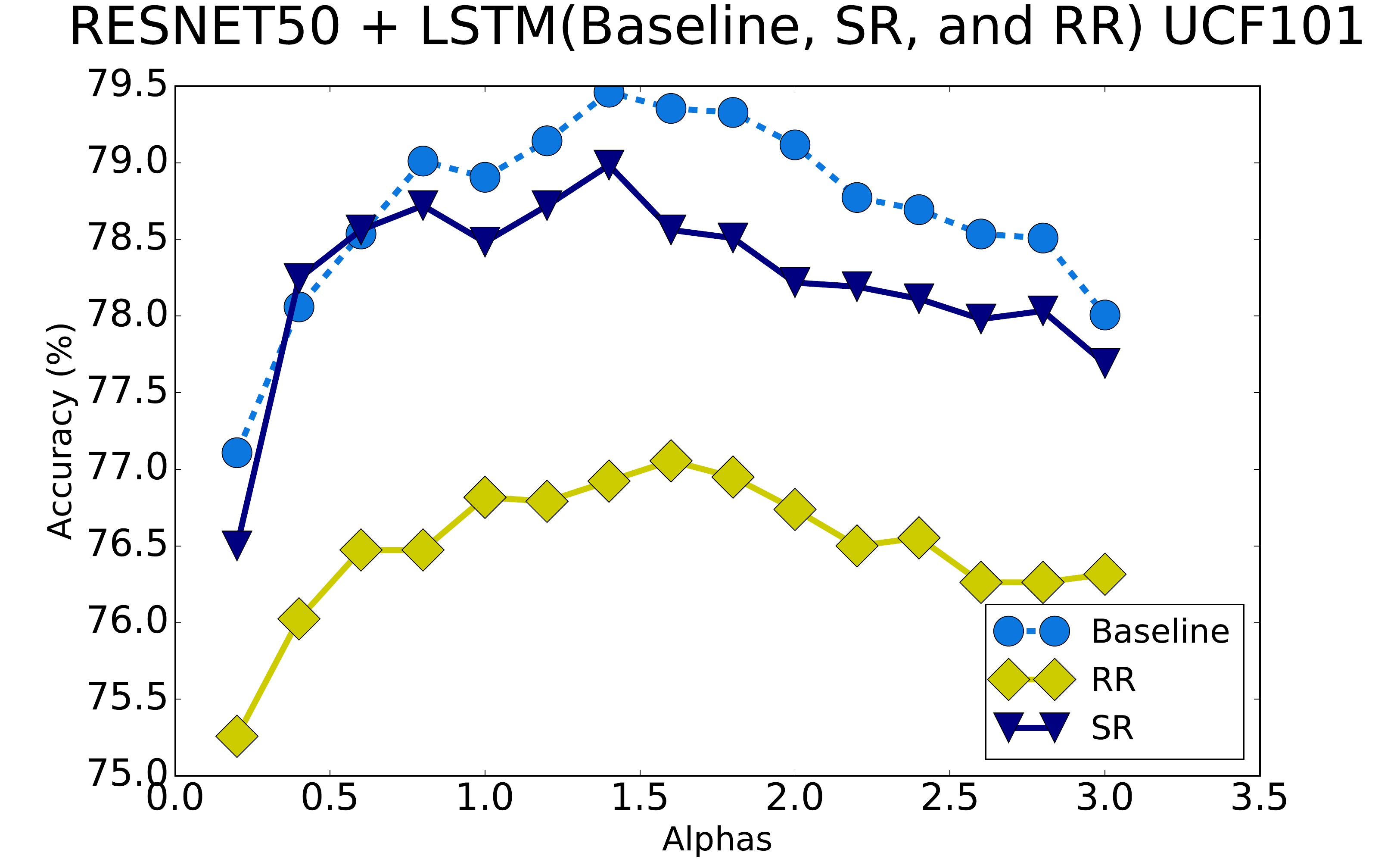}
  \end{subfigure}
  \caption{Input $\alpha$ testing results across HMDB51 and UCF101 for TSN and ResNet50 + LSTM are shown. Both these models show relatively high stability, which is apparent in the scale of accuracy values in the graph. For TSN model variants over HMDB51, there is minimal improvement in peak recognition accuracy with the baseline model showing the highest stability. 
However, when the same model is applied on UCF101, there is a distinct increase in the overall performance with SR-based TSN performing best. 
ResNet50 + LSTM shows the exact opposite trend, where RR and SR perform worse than the baseline when applied on UCF101 while on the HMDB51 data, it shows a significant improvement. 
The high stability combined with irregular behavior in peak recognition performance suggests an alternative temporal behavior pattern, which we label as the Type II category}
  \label{fig:tsn_conv_alpha_testing}
\end{figure}
Results from training models using SR and RR preprocessing are shown on the right-hand side of Fig.~\ref{fig:c3d_i3d_alpha_testing}.
Both preprocessing strategies result in a higher peak and a lower standard deviation in performance towards the extreme ends of the tested input $\alpha$ values when compared to the baseline.
I3D models trained using SR and RR preprocessing show similar stability to their C3D counterparts. Based on the similarities in recognition performance and stability, which characterizes a model's temporal behavior, we categorize C3D and I3D as Type I network architectures.

TSN and ConvNet + LSTM models show a different trend when compared to C3D or I3D models.
Fig.~\ref{fig:tsn_conv_alpha_testing} shows that the range of accuracies across these model variants is extremely low at $\sim 4\%$ as opposed to $\sim 18\%$ in Type I models.
This shows the inherently high stability of these models.
Dynamic \acro-based TSN models, when trained on HMDB51, show only a minor improvement in performance, and the baseline remains the most stable.
However, when trained on UCF101, we see a clear improvement in performance of dynamic \acro-based variants over the baseline across the entire range of input $\alpha$ values tested.
When comparing the results of ConvNet + LSTM to TSN variants, we see a reversal in the performance trends across datasets.
Here, HMDB51 results show an increase in performance of dynamic \acro-based models over the baseline across all input $\alpha$'s and UCF101 results show improved performance of the baseline model over most of the input $\alpha$'s.
From these results, we see that TSN and ConvNet + LSTM models show consistently high levels of stability across all model variants, and \acro-based models achieve higher peak recognition performance in most cases.
These behaviours are representative of Type II network architectures.
\section{Conclusion}

In this paper, we have shown that the problem of variation in speed of input signals to deep models is pertinent due to its detrimental impact on the performance of these models.
By assessing four state-of-the-art models in activity recognition, we have shown that the drop in performance could be up to $50.0\%$.
Through the use of \acro-based preprocessing, we address this issue and show a decrease in the worst-case drop in performance which is now only $6.7\%$.
Furthermore, through the use of dynamic \acro, we show an improvement in stability across all models by at least $4.7\%$. 
Based on our input $\alpha$ testing, we have identified two classes of models according to their temporal behavior.
Type I models exhibit low stability, which is improved by \acro-based models as well as reocgnition performance, while Type II models exhibit inherent stability, which is retained in \acro-based models along with a minor improvement in recognition performance.
We find that C3D and I3D behave as Type I models while TSN and ConvNet+LSTM behave as Type II models.

We believe that \acro-based preprocessing is critical for handling the variation in frame capture rate across videos in various datasets which is applicable to transfer learning.
We plan to extend this work by further analyzing the temporal behavior of Type II models by applying them to tasks with minimal visual semantic cues.

\section*{Acknowledgements}
Supported by the Intelligence Advanced Research Projects Activity (IARPA) via Department of Interior/ Interior Business Center (DOI/IBC) contract number D17PC00341.  The U.S. Government is authorized to reproduce and distribute reprints for Governmental purposes notwithstanding any copyright annotation thereon.  Disclaimer:  The views and conclusions contained herein are those of the authors and should not be interpreted as necessarily representing the official policies or endorsements, either expressed or implied, of IARPA, DOI/IBC, or the U.S. Government.

This work was also partially supported by NIST 60NANB17D191 and ARO W911NF-15-1-0354.

\clearpage

\bibliographystyle{splncs}
\bibliography{egbib}

\begin{thebibliography}{10}

\bibitem{laptev2005space}
Laptev, I.:
\newblock On space-time interest points.
\newblock International Journal of Computer Vision \textbf{64}(2-3) (2005)
  107--123

\bibitem{dalal2005histograms}
Dalal, N., Triggs, B.:
\newblock Histograms of oriented gradients for human detection.
\newblock In: 2005 IEEE Computer Society Conference on Computer Vision and
  Pattern Recognition (CVPR'05). Volume~1., IEEE (2005)  886--893

\bibitem{wang2011action}
Wang, H., Kl{\"a}ser, A., Schmid, C., Liu, C.L.:
\newblock Action recognition by dense trajectories.
\newblock In: Computer Vision and Pattern Recognition (CVPR), 2011 IEEE
  Conference on, IEEE (2011)  3169--3176

\bibitem{MyRaTASSP1981}
Myers, C.S., Rabiner, L.R., Rosenberg, A.E.:
\newblock Perfrmance tradeoffs in dynamic time warping algorithms for isolated
  word recognition.
\newblock IEEE Transactions on Acoustics, Speech, and Signal Processing
  \textbf{28}(6) (1980)  623--635

\bibitem{berndt1994using}
Berndt, D.J., Clifford, J.:
\newblock Using dynamic time warping to find patterns in time series.
\newblock In: KDD workshop. Volume~10., Seattle, WA (1994)  359--370

\bibitem{NIPS2014_5353}
Simonyan, K., Zisserman, A.:
\newblock Two-stream convolutional networks for action recognition in videos.
\newblock In Ghahramani, Z., Welling, M., Cortes, C., Lawrence, N.D.,
  Weinberger, K.Q., eds.: Advances in Neural Information Processing Systems 27.
\newblock Curran Associates, Inc. (2014)  568--576

\bibitem{tran2015learning}
Tran, D., Bourdev, L., Fergus, R., Torresani, L., Paluri, M.:
\newblock Learning spatiotemporal features with 3d convolutional networks.
\newblock In: 2015 IEEE International Conference on Computer Vision (ICCV),
  IEEE (2015)  4489--4497

\bibitem{wang2015temporal}
Wang, P., Cao, Y., Shen, C., Liu, L., Shen, H.T.:
\newblock Temporal pyramid pooling based convolutional neural networks for
  action recognition.
\newblock arXiv preprint arXiv:1503.01224 (2015)

\bibitem{feichtenhofer2016convolutional}
Feichtenhofer, C., Pinz, A., Zisserman, A.:
\newblock Convolutional two-stream network fusion for video action recognition.
\newblock arXiv preprint arXiv:1604.06573 (2016)

\bibitem{hochreiter1997long}
Hochreiter, S., Schmidhuber, J.:
\newblock Long short-term memory.
\newblock Neural computation \textbf{9}(8) (1997)  1735--1780

\bibitem{cho2014properties}
Cho, K., Van~Merri{\"e}nboer, B., Bahdanau, D., Bengio, Y.:
\newblock On the properties of neural machine translation: Encoder-decoder
  approaches.
\newblock arXiv preprint arXiv:1409.1259 (2014)

\bibitem{donahue2015long}
Donahue, J., Anne~Hendricks, L., Guadarrama, S., Rohrbach, M., Venugopalan, S.,
  Saenko, K., Darrell, T.:
\newblock Long-term recurrent convolutional networks for visual recognition and
  description.
\newblock In: Proceedings of the IEEE Conference on Computer Vision and Pattern
  Recognition. (2015)  2625--2634

\bibitem{kay2017kinetics}
Kay, W., Carreira, J., Simonyan, K., Zhang, B., Hillier, C., Vijayanarasimhan,
  S., Viola, F., Green, T., Back, T., Natsev, P.,  et~al.:
\newblock The kinetics human action video dataset.
\newblock arXiv preprint arXiv:1705.06950 (2017)

\bibitem{wang2016temporal}
Wang, L., Xiong, Y., Wang, Z., Qiao, Y., Lin, D., Tang, X., Van~Gool, L.:
\newblock Temporal segment networks: Towards good practices for deep action
  recognition.
\newblock In: European Conference on Computer Vision, Springer (2016)  20--36

\bibitem{klaser2008spatio}
Klaser, A., Marsza{\l}ek, M., Schmid, C.:
\newblock A spatio-temporal descriptor based on 3d-gradients.
\newblock In: BMVC 2008-19th British Machine Vision Conference, British Machine
  Vision Association (2008)  275--1

\bibitem{liao2010region}
Liao, W.H.:
\newblock Region description using extended local ternary patterns.
\newblock In: Pattern Recognition (ICPR), 2010 20th International Conference
  on, IEEE (2010)  1003--1006

\bibitem{willems2008efficient}
Willems, G., Tuytelaars, T., Van~Gool, L.:
\newblock An efficient dense and scale-invariant spatio-temporal interest point
  detector.
\newblock In: European conference on computer vision, Springer (2008)  650--663

\bibitem{raptis2010tracklet}
Raptis, M., Soatto, S.:
\newblock Tracklet descriptors for action modeling and video analysis.
\newblock In: European conference on computer vision, Springer (2010)  577--590

\bibitem{lu2010learning}
Lu, W.C., Wang, Y.C.F., Chen, C.S.:
\newblock Learning dense optical-flow trajectory patterns for video object
  extraction.
\newblock In: Advanced Video and Signal Based Surveillance (AVSS), 2010 Seventh
  IEEE International Conference on, IEEE (2010)  315--322

\bibitem{simonyan2014two}
Simonyan, K., Zisserman, A.:
\newblock Two-stream convolutional networks for action recognition in videos.
\newblock In: Advances in neural information processing systems. (2014)
  568--576

\bibitem{6165309}
Ji, S., Xu, W., Yang, M., Yu, K.:
\newblock 3d convolutional neural networks for human action recognition.
\newblock IEEE Transactions on Pattern Analysis and Machine Intelligence
  \textbf{35}(1) (Jan 2013)  221--231

\bibitem{sadanand2012action}
Sadanand, S., Corso, J.J.:
\newblock Action bank: A high-level representation of activity in video.
\newblock In: Computer Vision and Pattern Recognition (CVPR), 2012 IEEE
  Conference on, IEEE (2012)  1234--1241

\bibitem{hara2017learning}
Hara, K., Kataoka, H., Satoh, Y.:
\newblock Learning spatio-temporal features with 3d residual networks for
  action recognition.
\newblock In: Proceedings of the ICCV Workshop on Action, Gesture, and Emotion
  Recognition. (2017) ~4

\bibitem{carreira2017quo}
Carreira, J., Zisserman, A.:
\newblock Quo vadis, action recognition? a new model and the kinetics dataset.
\newblock In: 2017 IEEE Conference on Computer Vision and Pattern Recognition
  (CVPR), IEEE (2017)  4724--4733

\bibitem{deng2009imagenet}
Deng, J., Dong, W., Socher, R., Li, L.J., Li, K., Fei-Fei, L.:
\newblock Imagenet: A large-scale hierarchical image database.
\newblock In: Computer Vision and Pattern Recognition, 2009. CVPR 2009. IEEE
  Conference on, IEEE (2009)  248--255

\bibitem{piergiovanni2017learning}
Piergiovanni, A., Fan, C., Ryoo, M.S.:
\newblock Learning latent sub-events in activity videos using temporal
  attention filters.
\newblock In: Proceedings of the 31st AAAI conference on artificial
  intelligence. (2017)

\bibitem{tallec2018can}
Tallec, C., Ollivier, Y.:
\newblock Can recurrent neural networks warp time?
\newblock In: International Conference on Learning Representations. (2018)

\bibitem{kuehne2013hmdb51}
Kuehne, H., Jhuang, H., Stiefelhagen, R., Serre, T.:
\newblock Hmdb51: A large video database for human motion recognition.
\newblock In: High Performance Computing in Science and Engineering `12.
\newblock Springer (2013)  571--582

\bibitem{soomro2012ucf101}
Soomro, K., Zamir, A.R., Shah, M.:
\newblock Ucf101: A dataset of 101 human actions classes from videos in the
  wild.
\newblock arXiv preprint arXiv:1212.0402 (2012)

\bibitem{ioffe2015batch}
Ioffe, S., Szegedy, C.:
\newblock Batch normalization: Accelerating deep network training by reducing
  internal covariate shift.
\newblock In: International conference on machine learning. (2015)  448--456

\end{thebibliography}

\newpage
\appendix 
\section{Preprocessing Steps}
\label{app:setup}
The following section details the preprocessing steps and experimental setup applied to each model.

\subsection{C3D}
The preprocessing steps applied to the inputs provided to the C3D model are,
\begin{itemize}
\item Extract a clip of 50 frames from a given input video, using a random offset. If the video is less than 50 frames, loop the video to obtain atleast 50 frames.
\item Resize image, preserving the aspect ratio, to a minimum side of 112.
\item Crop the central portion of the frame to fit $112 \times 112$ dimensions.
\item Subtract the mean image block calculated for 16 frames across the entire dataset.
\end{itemize}

\subsection{I3D}
The preprocessing steps applied to the inputs provided to the I3D model are,
\begin{itemize}
\item Extract a clip of 250 frames from a given input video, using random offset. If the video is less than 250 frames, loop the video to obtain atleast 250 frames.
\item Resize image, preserving the aspect ratio, to a minimum side of 256.
\item Crop the central portion of the frame to fit $224 \times 224$ dimensions.
\item Rescale the image values between [-1, 1].
\end{itemize}

\subsection{TSN}
The preprocessing steps applied to the inputs provided to the TSN model are,
\begin{itemize}
\item Reduce a given video to 180 frames (3 segments of 60 frames each) by uniformly sampling across the video.
\item Resize the frame to $224 \times 224$ dimensions.
\item Rescale and crop each segment, using a fixed aspect ratio, with minimum side 256, .
\item Apply random horizontal flipping to each segment.
\item Subtract the mean RGB values, R = 123, G = 117 and B = 104.
\end{itemize}

\subsection{ResNet50 + LSTM}
The preprocessing steps applied to the inputs provided to the ConvNet + LSTM model are,
\begin{itemize}
\item Extract a clip of 250 frames from a given input video, using a random offset. If the video is less than 250 frames, loop the video to obtain atleast 250 frames.
\item Resize image, preserving the aspect ratio, to a minimum side of 256.
\item Crop the central portion of the frame to fit $224 \times 224$ dimensions.
\item Subtract the mean RGB values, R = 123.68, G = 116.78 and B = 103.94.
\end{itemize}
\hfill \\
\section{Extended results for evaluation of temporal robustness}
\label{app:ext_res}
Tables~\ref{app:extended_results_1} and~\ref{app:extended_results_2} show the effects of modifying speed of input videos on the performance of each of the baseline models. 
In certain exaggerated cases, the difference in recognition accuracy can rise up to $50\%$, as in C3D for the action class ``'flic-flac''. 
The extended set of tables is intended to supplement the motivating factor of this work, to protect deep models from target adversarial attacks based on variation in speed of input.
An additional point to note is that the trend in maximum difference of performance across different models supports the argument for typical characteristics of Type I and Type II network architectures.
\begin{table}[th!]
\centering
\caption{Evaluation results of baseline models on the HMDB51Rate dataset are shown. Videos are divided into four bins based on resampling factors, $[0.2-0.6], [0.6-1.0],[1.0-2.0], \text{and} [2.0-3.0]$. M.A. refers to mean action class accuracy across the four bins while M.D. refers to the maximum difference in recognition accuracy between each of the four bins}
\label{app:extended_results_1}
\begin{tabular}{l|c|c|c|c|c|c|c|c}
\hline
\multirow{2}{*}{Action Class}&\multicolumn{2}{|c}{C3D}&\multicolumn{2}{|c}{I3D}&\multicolumn{2}{|c}{TSN}&\multicolumn{2}{|c}{ResNet50+LSTM}\\\cline{2-9}
\multicolumn{1}{c}{}&\multicolumn{1}{|c|}{M.A.}&\multicolumn{1}{c|}{M.D.}&\multicolumn{1}{c|}{M.A.}&\multicolumn{1}{c|}{M.D.}&\multicolumn{1}{c|}{M.A.}&\multicolumn{1}{c|}{M.D.}&\multicolumn{1}{c|}{M.A.}&\multicolumn{1}{c}{M.D.}\\
\hline
\hline
brush\_hair      & 67.8 &  6.67 & 76.9  &  6.67 & 46.9  &  6.67 & 39.6  & 10.0 \\
cartwheel        & 14.0 & 33.3  & 27.6  & 43.3  &  8.67 &  6.67 & 23.8  &  6.67\\
catch            & 80.0 & 10.0  & 56.2  & 16.7  & 43.3  & 13.3  & 50.2  &  3.33\\
chew             & 22.4 & 13.3  & 64.0  & 46.7  & 25.3  & 10.0  & 56.0  &  6.67\\
clap             & 10.4 & 13.3  & 50.0  & 30.0  & 50.2  &  6.67 & 42.9  & 13.3 \\
climb            & 77.6 & 16.7  & 91.1  & 16.7  & 41.6  & 10.0  & 64.7  &  6.67\\
climb\_stairs    & 43.1 & 20.0  & 68.0  & 20.0  & 50.2  & 23.3  & 39.3  &  6.67\\
dive             & 60.9 & 33.3  & 55.6  & 16.7  & 51.6  & 16.7  & 82.4  &  3.33\\
draw\_sword      & 27.8 & 16.7  & 39.3  & 20.0  & 51.8  &  3.33 & 32.9  &  3.33\\
dribble          & 72.0 & 13.3  & 96.4  &  6.67 & 72.4  &  6.67 & 70.0  &  0.00\\
drink            & 15.3 & 10.0  & 58.7  & 23.3  & 42.2  &  6.67 & 46.7  & 10.0 \\
eat              & 29.1 &  6.67 & 57.8  & 20.0  & 47.8  &  6.67 & 46.4  & 13.3 \\
fall\_floor      & 20.7 & 26.7  & 22.0  & 33.3  & 29.8  &  3.33 & 28.7  & 20.0 \\
fencing          & 41.3 & 16.7  & 57.1  & 30.0  & 57.6  &  6.67 & 41.6  &  6.67\\
flic\_flac       & 28.2 & 50.0  & 36.4  & 40.0  & 19.6  & 13.3  & 44.4  & 16.7 \\
golf             & 93.8 &  3.33 & 99.6  &  6.67 & 78.7  & 13.3  & 88.9  &  6.67\\
handstand        & 70.7 & 13.3  & 92.2  & 30.0  & 31.3  &  6.67 & 53.6  & 10.0 \\
hit              & 24.2 & 16.7  & 31.3  & 33.3  &  6.89 &  3.33 & 32.2  & 13.3 \\
hug              & 46.2 & 33.3  & 58.9  & 43.3  & 69.1  & 16.7  & 55.8  & 16.7 \\
jump             & 39.6 & 23.3  & 42.4  & 50.0  & 26.9  & 10.0  & 25.6  & 10.0 \\
kick             & 33.6 & 40.0  &  2.89 & 10.0  & 11.1  &  3.33 & 12.0  &  6.67\\
kick\_ball       & 39.3 & 30.0  & 46.2  & 46.7  & 38.9  & 13.3  & 40.9  & 10.0 \\
kiss             & 79.8 & 13.3  & 82.0  & 13.3  & 60.7  &  3.33 & 78.4  &  3.33\\
laugh            & 25.1 &  6.67 & 73.1  & 40.0  & 37.1  & 10.0  & 25.3  & 10.0 \\
pick             & 23.8 & 16.7  & 44.7  & 26.7  &  2.2  &  3.33 &  8.89 &  6.67\\
\hline
\end{tabular}
\end{table}

\begin{table}[t!]
\centering
\caption{Evaluation results of baseline models on the HMDB51Rate dataset are shown. Videos are divided into four bins based on resampling factors, $[0.2-0.6], [0.6-1.0],[1.0-2.0], \text{and} [2.0-3.0]$. M.A. refers to mean action class accuracy across the four bins while M.D. refers to the maximum difference in recognition accuracy between each of the four bins. There network architectures are extremely susceptible to targeted attacks on each class, with variation in speed of inputs affecting performance directly}
\label{app:extended_results_2}
\begin{tabular}{l|c|c|c|c|c|c|c|c}
\hline
\multirow{2}{*}{Action Class}&\multicolumn{2}{|c}{C3D}&\multicolumn{2}{|c}{I3D}&\multicolumn{2}{|c}{TSN}&\multicolumn{2}{|c}{ResNet50+LSTM}\\\cline{2-9}
\multicolumn{1}{c}{}&\multicolumn{1}{|c|}{M.A.}&\multicolumn{1}{c|}{M.D.}&\multicolumn{1}{c|}{M.A.}&\multicolumn{1}{c|}{M.D.}&\multicolumn{1}{c|}{M.A.}&\multicolumn{1}{c|}{M.D.}&\multicolumn{1}{c|}{M.A.}&\multicolumn{1}{c}{M.D.}\\
\hline
\hline
pour             & 62.0 &  6.67 & 85.1  &  3.33 & 70.0  & 16.7  & 60.4  &  3.33\\
pullup           & 92.4 & 16.7  & 92.7  & 16.7  & 100.0 & 0.00  & 81.6  &  3.33\\
punch            & 20.4 & 43.3  & 44.0  & 26.7  &  9.78 & 10.0  & 38.0  & 16.7 \\
push             & 56.9 & 36.7  & 79.3  & 13.3  & 35.6  & 10.0  & 76.2  & 10.0 \\
pushup           & 84.9 & 26.7  & 81.8  & 26.7  & 83.6  &  3.33 & 57.8  & 10.0 \\
ride\_bike       & 90.9 &  6.67 & 99.6  &  3.33 & 95.1  & 10.0  & 94.4  &  3.33\\
ride\_horse      & 79.1 & 10.0  & 89.3  & 16.7  & 55.6  & 16.7  & 70.2  & 16.7 \\
run              & 39.6 & 23.3  & 16.0  & 20.0  & 57.3  & 10.0  & 43.1  & 10.0 \\
shake\_hands     & 50.9 & 23.3  & 80.7  & 20.0  & 70.4  & 10.0  & 62.0  & 10.0 \\
shoot\_ball      & 93.3 &  6.67 & 87.1  & 16.7  & 43.6  & 13.3  & 63.8  &  6.67\\
shoot\_bow       & 90.7 & 13.3  & 91.8  & 10.0  & 88.9  &  3.33 & 75.1  &  3.33\\
shoot\_gun       & 63.8 & 13.3  & 83.8  & 16.7  & 70.9  &  3.33 & 68.0  &  3.33\\
sit              & 24.7  & 20.0  & 20.7  & 16.7  & 43.3  & 13.3  & 22.9  & 10.0 \\
situp            & 95.8  &  6.67 & 100.0 &  0.00 & 94.0  & 10.0  & 75.8  & 13.3 \\
smile            & 38.0  & 20.0  & 46.0  & 16.7  & 33.3  &  6.67 & 35.1  &  3.33\\
smoke            & 20.2  & 10.0  & 79.8  & 10.0  & 17.1  & 20.0  & 49.3  & 13.3 \\
somersault       & 48.7  & 13.3  & 56.9  & 10.0  & 13.6  &  6.67 & 53.6  & 16.7 \\
stand            & 18.7  & 26.7  & 14.2  & 16.7  & 17.8  &  6.67 &  6.89 &  6.67\\
swing\_baseball  & 14.0  &  6.67 & 40.0  & 26.7  & 16.2  & 10.0  &  6.22 &  3.33\\
sword            & 11.6  & 10.0  & 16.4  & 23.3  & 23.3  &  6.67 & 14.2  &  6.67\\
sword\_exercise  &  5.33 &  6.67 & 28.4  & 20.0  & 10.0  &  0.00 & 15.6  &  6.67\\
talk             & 40.9  & 10.0  & 80.9  & 10.0  & 53.3  &  6.67 & 26.7  &  6.67\\
throw            & 25.8  & 13.3  & 41.3  & 23.3  & 51.3  &  6.67 & 11.3  &  3.33\\
turn             & 23.6  & 20.0  & 30.4  & 30.0  &  3.78 &  6.67 & 40.7  &  6.67\\
walk             & 21.1  & 10.0  & 44.2  & 23.3  & 25.8  &  6.67 & 17.8  & 13.3 \\
wave             &  2.00 & 10.0  &  8.89 & 10.0  & 12.7  &  6.67 & 12.4  & 10.0 \\
\hline
\hline
\textbf{Total Average}   & \textbf{45.13} & \textbf{17.52} & \textbf{58.23} & \textbf{21.50} & \textbf{43.10} &  \textbf{8.69} & \textbf{45.30} &  \textbf{8.56}\\
\hline
\end{tabular}
\end{table}

\clearpage
\newpage
\section{Additional analysis from stability plots}
\begin{figure}[h!]
  \begin{subfigure}[h]{0.49\textwidth}
    \centering
    \includegraphics[width=\textwidth]{C3D_CVR_alpha.pdf}
  \end{subfigure}
  \begin{subfigure}[h]{0.49\textwidth}
    \centering
    \includegraphics[width=\textwidth]{C3D_SR_RR_alpha.pdf}
  \end{subfigure}
  \begin{subfigure}[h]{0.49\textwidth}
    \centering
    \includegraphics[width=\textwidth]{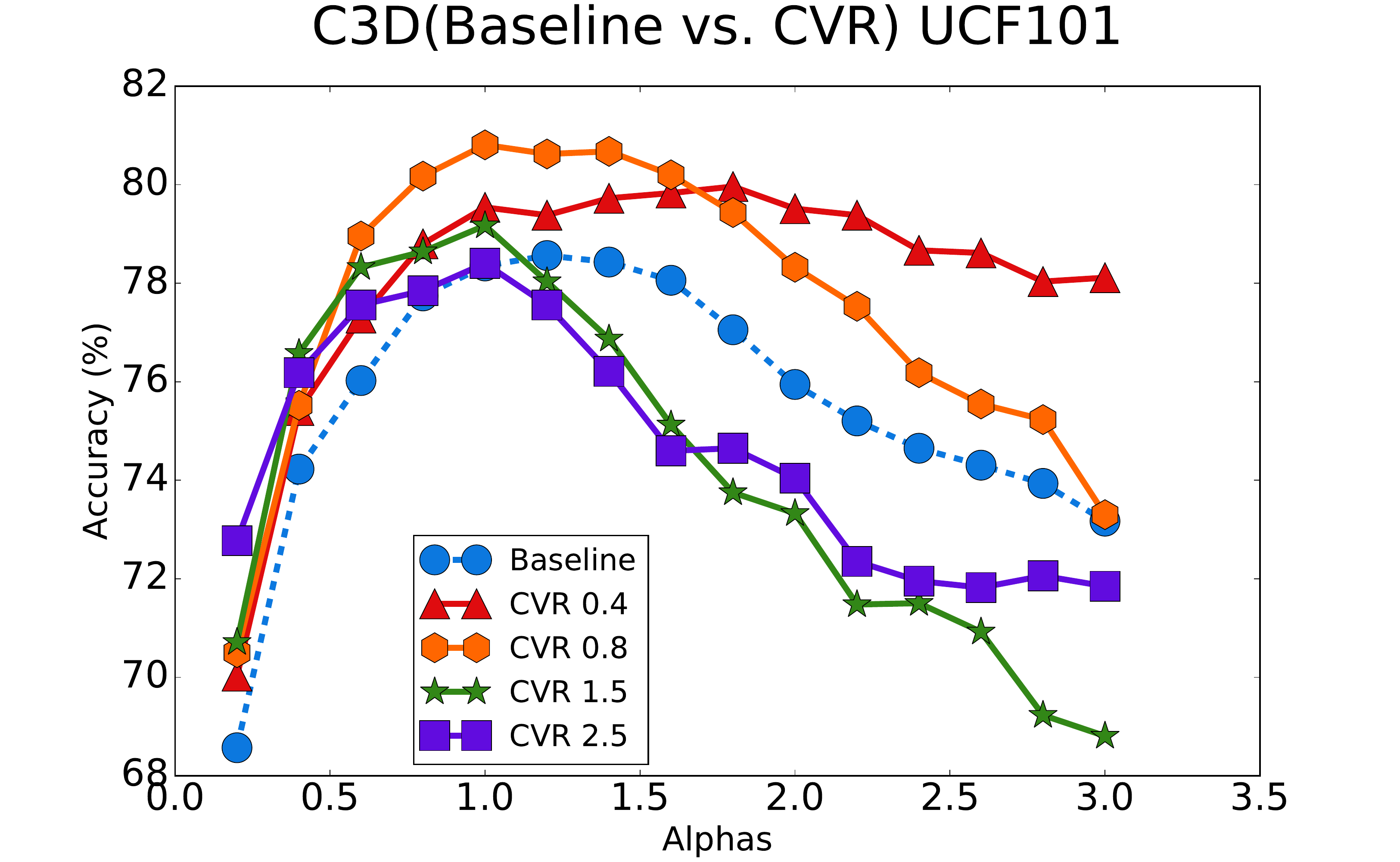}
  \end{subfigure}
  \begin{subfigure}[h]{0.49\textwidth}
    \centering
    \includegraphics[width=\textwidth]{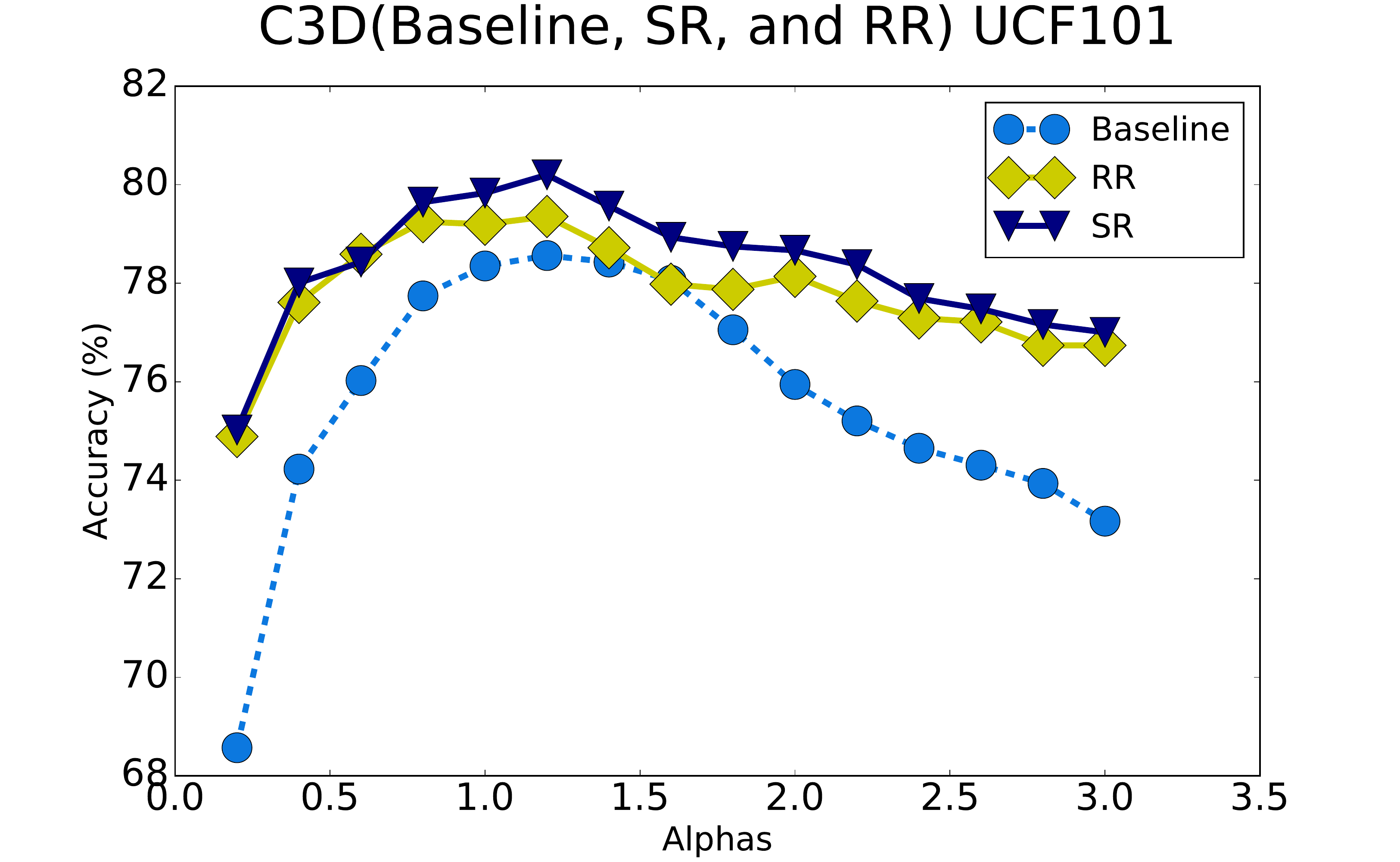}
  \end{subfigure}
  \begin{subfigure}[h]{0.49\textwidth}
    \centering
    \includegraphics[width=\textwidth]{I3D_CVR_alpha.pdf}
  \end{subfigure}
  \begin{subfigure}[h]{0.49\textwidth}
    \centering
    \includegraphics[width=\textwidth]{I3D_SR_RR_alpha.pdf}
  \end{subfigure}
  \begin{subfigure}[h]{0.49\textwidth}
    \centering
    \includegraphics[width=\textwidth]{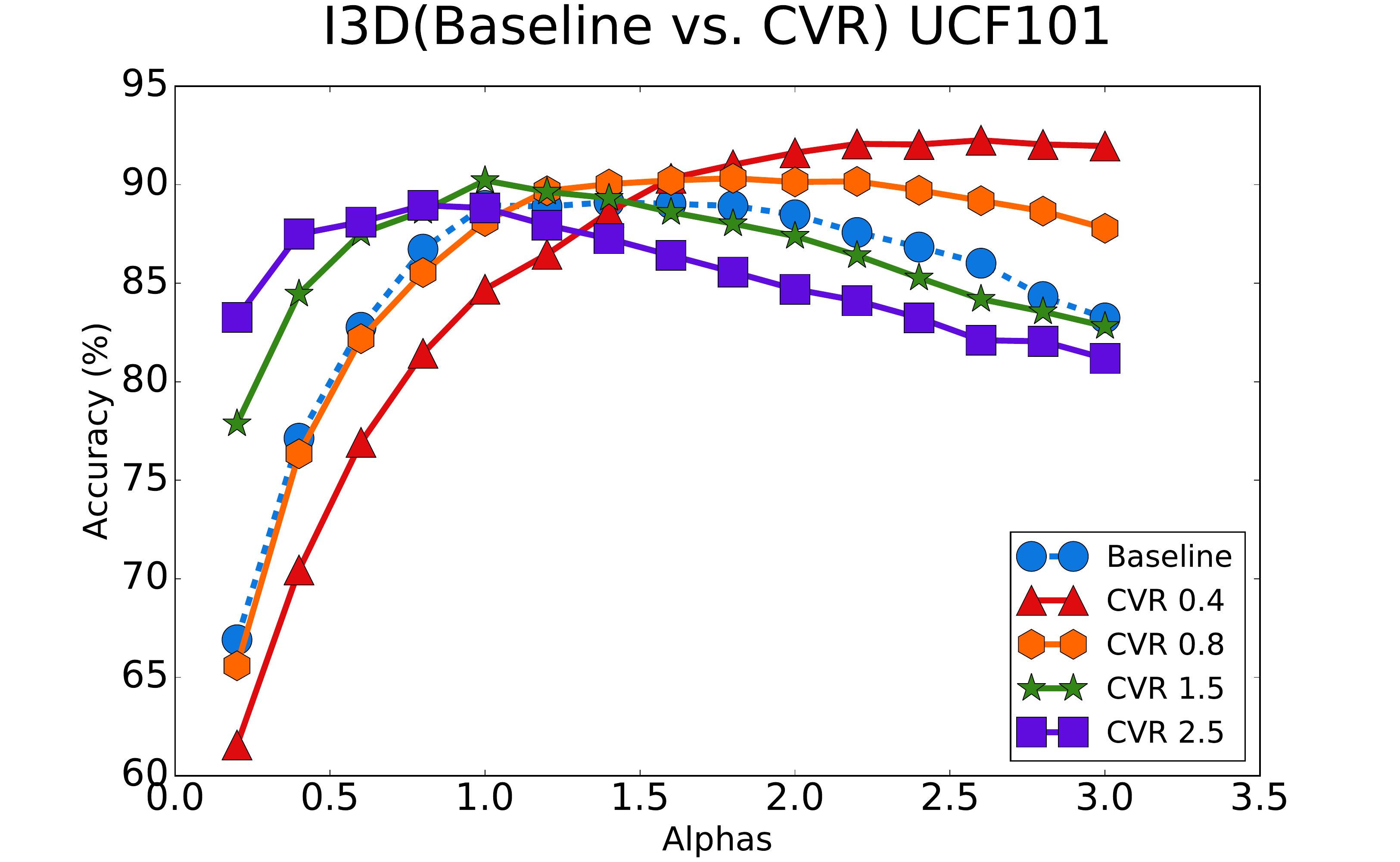}
  \end{subfigure}
  \begin{subfigure}[h]{0.49\textwidth}
    \centering
    \includegraphics[width=\textwidth]{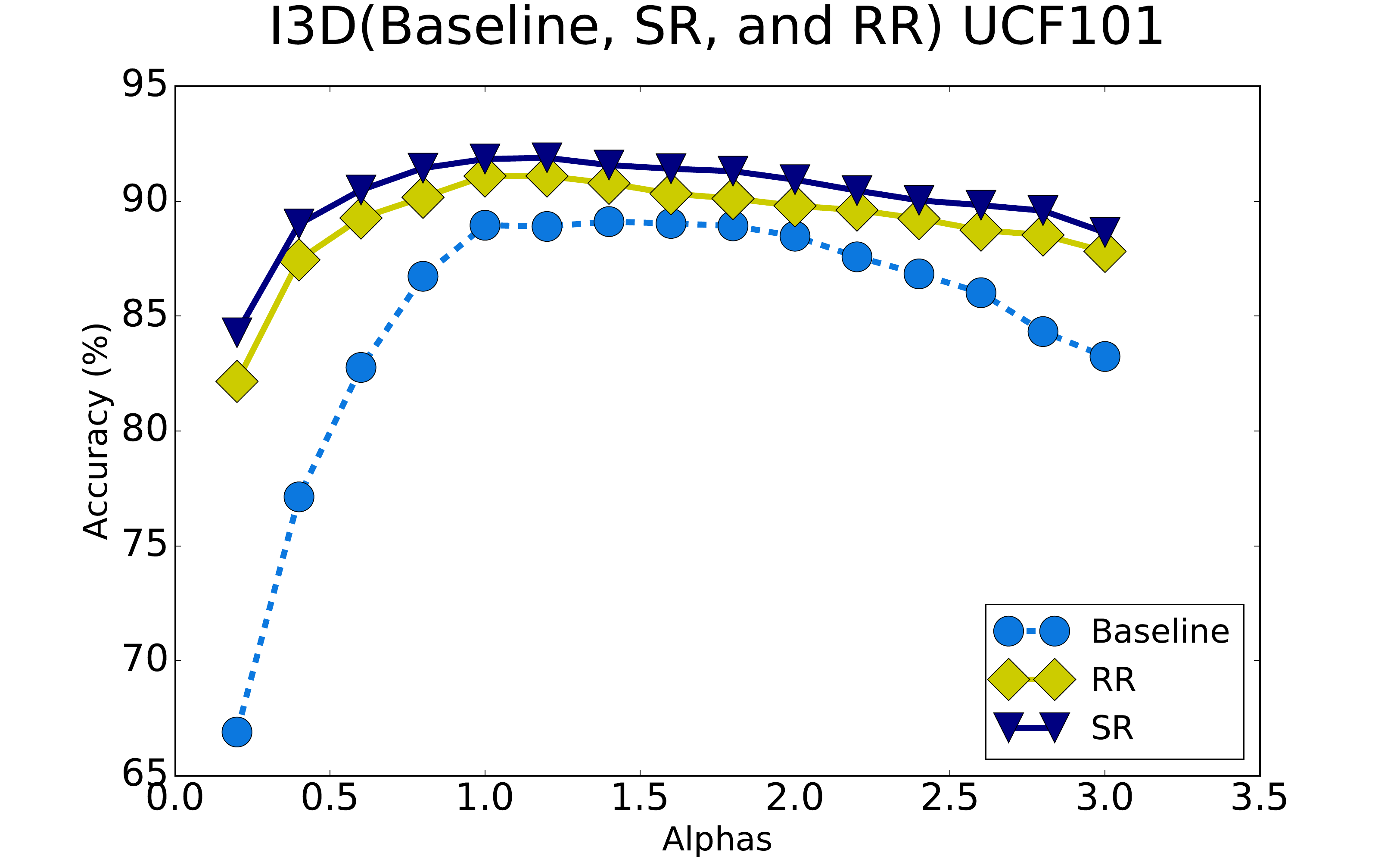}
  \end{subfigure}
  \caption{Stability in performance across input $\alpha$ tests is shown for C3D and I3D models. There models show a distinct drop in performance at the extremeties of speed when compared to Type II models since there is minimal change in information within a small block of neighboring frames due to sever under/over-sampling. Type I models operate on spatio-temporal features as opposed to frame level features, as in the case of Type II models, thus accounting for the drop in performance.}
  \label{fig:c3d_i3d_complete_alpha_testing}
\end{figure}

\begin{figure}[h]
  \begin{subfigure}[h]{0.49\textwidth}
    \centering
    \includegraphics[width=\textwidth]{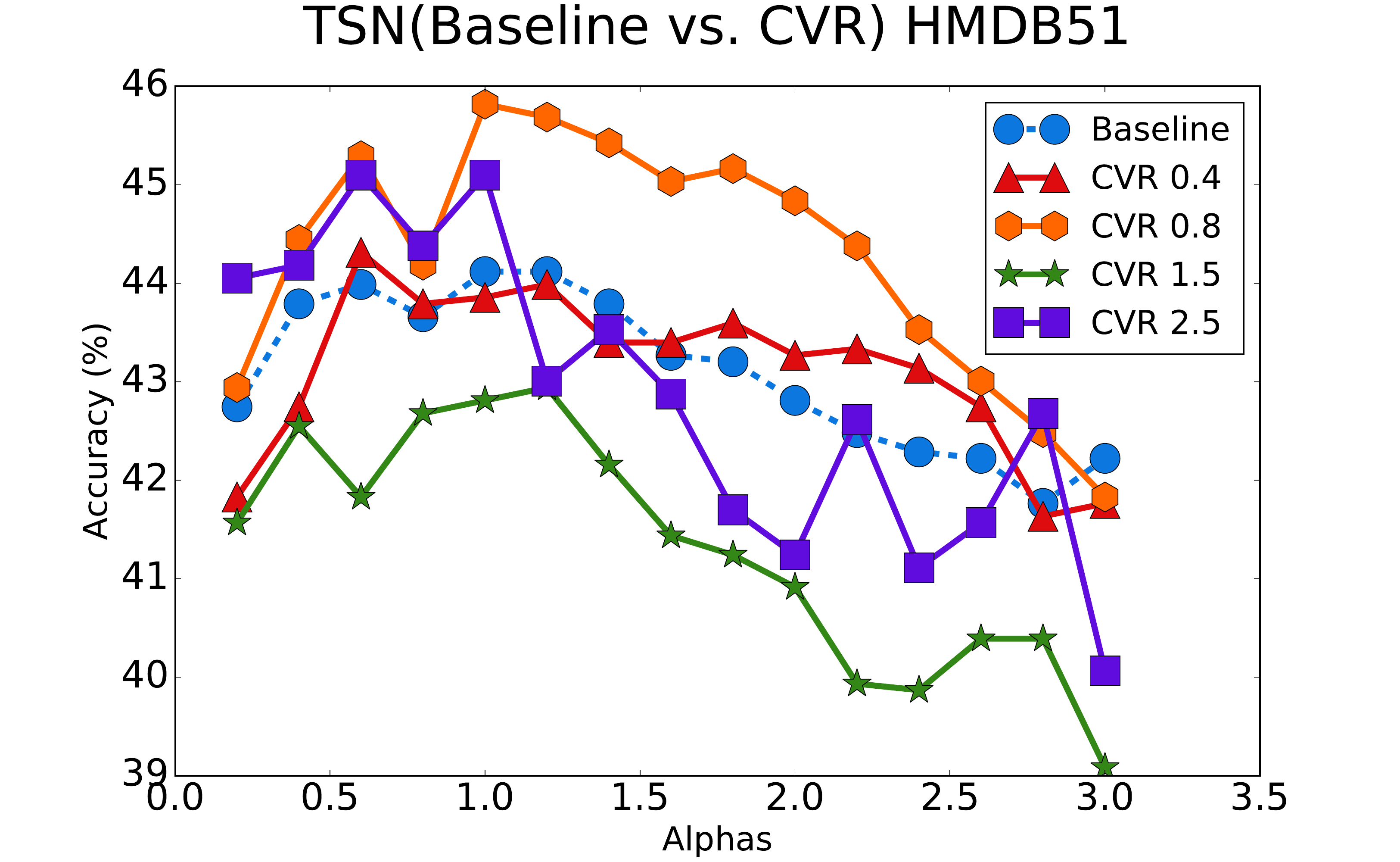}
  \end{subfigure}
  \begin{subfigure}[h]{0.49\textwidth}
    \centering
    \includegraphics[width=\textwidth]{TSN_SR_RR_alpha.pdf}
  \end{subfigure}
  \begin{subfigure}[h]{0.49\textwidth}
    \centering
    \includegraphics[width=\textwidth]{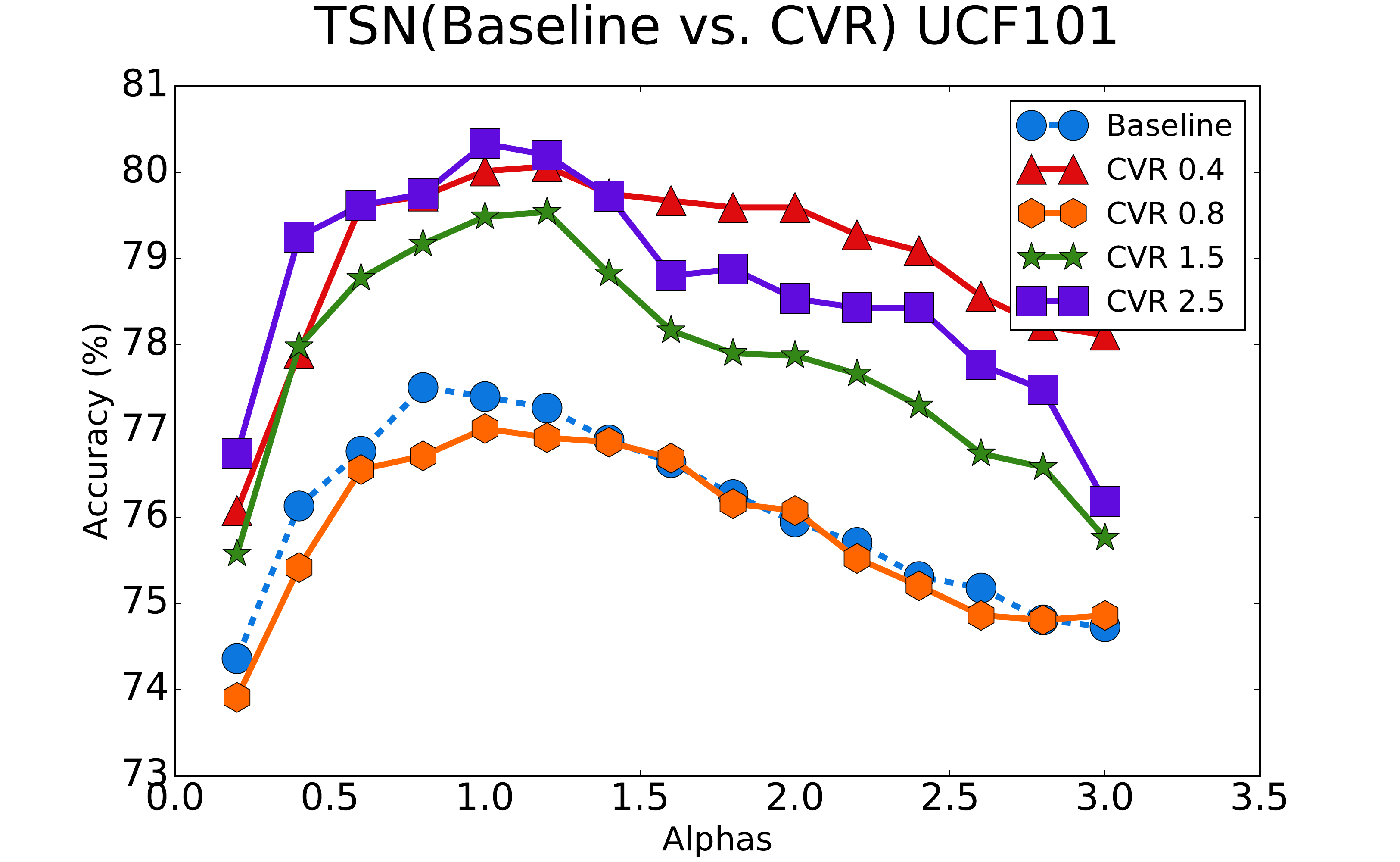}
  \end{subfigure}
  \begin{subfigure}[h]{0.49\textwidth}
    \centering
    \includegraphics[width=\textwidth]{TSN_SR_RR_UCF101.pdf}
  \end{subfigure}
  \begin{subfigure}[h]{0.49\textwidth}
    \centering
    \includegraphics[width=\textwidth]{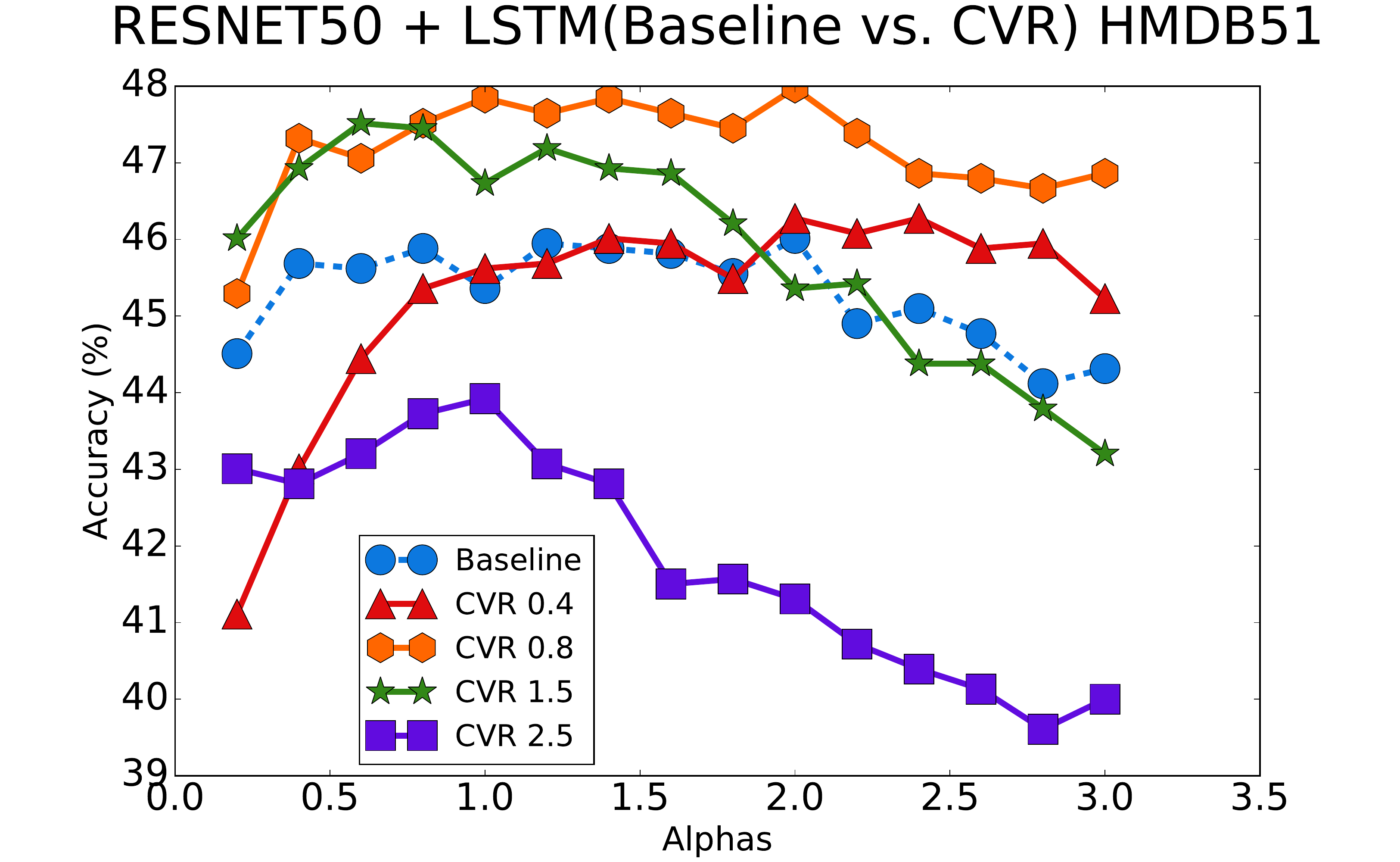}
  \end{subfigure}
  \begin{subfigure}[h]{0.49\textwidth}
    \centering
    \includegraphics[width=\textwidth]{Resnet_SR_RR_alpha.pdf}
  \end{subfigure}
  \begin{subfigure}[h]{0.49\textwidth}
    \centering
    \includegraphics[width=\textwidth]{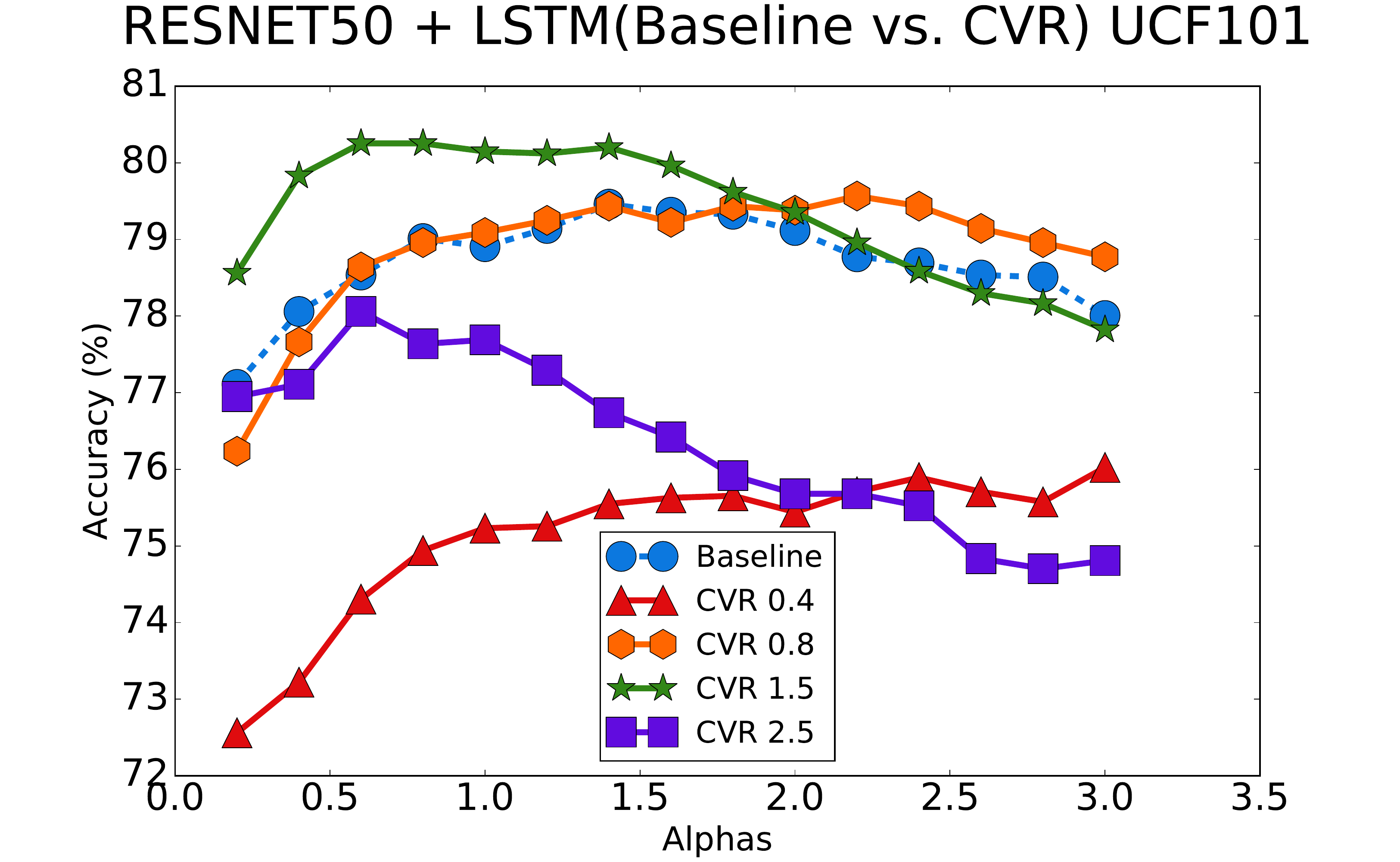}
  \end{subfigure}
  \begin{subfigure}[h]{0.49\textwidth}
    \centering
    \includegraphics[width=\textwidth]{Resnet_SR_RR_UCF101.pdf}
  \end{subfigure}
  \caption{Within the performance stability plots for TSN and ResNet50 + LSTM models, there is a distinct difference in performance levels between SR and RR model variants on UCF101 while varying minimally on the HMDB51 dataset. This can be correlated to the size of the dataset, where for smaller datasets, RR exposes the model to the entire range of possible speeds, which is more important to learn robust models. For larger datasets, we assume an inherent capacity to model the general speed variations and hence SR exposes models to more extreme variations in order to achieve peak performance.}
  \label{fig:tsn_resnet_complete_alpha_testing}
\end{figure}

From Figs.~\ref{fig:c3d_i3d_complete_alpha_testing} and~\ref{fig:tsn_resnet_complete_alpha_testing}, one easily observable phenomenon is the sharp drop in recognition performance at extremely high or low sampling factors (speed).
The primary reason for this phenomenon is that at such resampling factors, frames get repeated a number of times and there is no significant change in information within their immediately neighboring set.
Given that C3D and I3D act upon spatiotemporal blocks of information, minimal change within a block has a detrimental effect on their performance.

Comparing this to Fig.~\ref{fig:tsn_resnet_complete_alpha_testing}, Type II models, by construction, operate on frame level features using some form of temporal modeling, ``consensus'' in TSN and LSTM in ResNet50 + LSTM.
The last frame's prediction is often the most influential in case of ConvNet + LSTM models while TSN operates at an extremely high abstract level when associating individual frames to an action.
Thus, they do not offer as drastic a drop when compared to C3D or I3D.
From the right-hand side of Fig.~\ref{fig:c3d_i3d_complete_alpha_testing}, we see that SR and RR models help provide more robustness to C3D and I3D models by exposing them to more extreme variations within the fixed temporal window used to process information. 
Thus, they are able to reduce the severity of the drop in performance.

Another interesting trend observable from the right-hand side graphs of Fig.~\ref{fig:tsn_resnet_complete_alpha_testing} is the distinct difference in performance and stability of Type II SR and RR models across datasets.
Here, for HMDB51 SR and RR variants seems to perform comparably to each other while on UCF101, there is a clear difference in performance levels.
\begin{figure}[t!]
\centering
\includegraphics[width=0.8\textwidth]{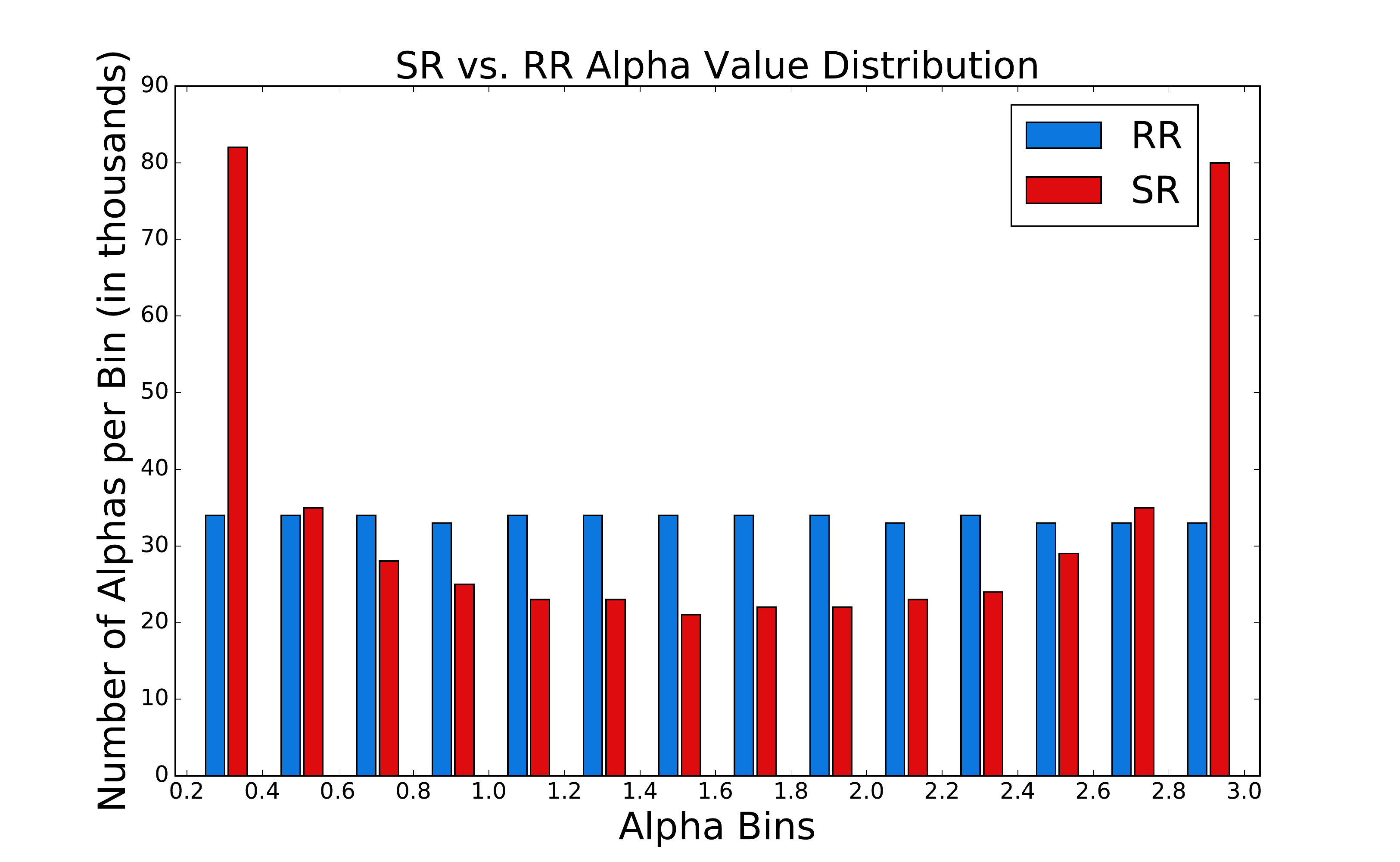}
\caption{The difference in SR and RR model performances varies in part due to the differences in the alpha values that get selected during the training regime. RR samples from a uniform distribution leading to similar numbers of alphas being selected form each alpha bin. SR samples alpha values according to the current number of videos loaded. The structure of a sine wave causes the extreme values of the distribution to be sampled more often in SR. More specifically this means that SR models are most likely to see $\alpha$ values between $0.2 \text{ to } 0.4$ and $2.8 \text{ to } 3.0$ where the sampling algorithm begins to cause videos to consist primarily of repeated frames. This plot consists of $476,848$ $\alpha$ values used while training C3D on UCF101 for $10$ epochs.}
\label{fig:sr_rr_dist}
\end{figure}
Using Fig.~\ref{fig:sr_rr_dist}, we observe that in the case of RR, the model is exposed to a uniform set of possible variations across the entire range of speeds within a dataset while in SR, there is a higher chance of being exposed to extreme variations.
We believe that Type II network architectures' performance have a direct dependency on the size of dataset being processed. 
For smaller datasets like HMDB51, RR exposes the model to the general set of possible speed variations in input which is equally if not more critical to improve the performance of the model. 
However, for larger datasets, that implicitly cover the so called ``middle'' range of speed variations, to extract maximum performance, the focus on exposing the model to extreme variations using SR allows for better performance.
Extending this argument, C3D and I3D are pretrained on large video datasets and hence show consistently high improvement in performance when using SR.

\clearpage

\section{Checkpoint Testing}
\begin{figure}[h!]
  \begin{subfigure}[h]{0.49\textwidth}
    \centering
    \includegraphics[width=\textwidth]{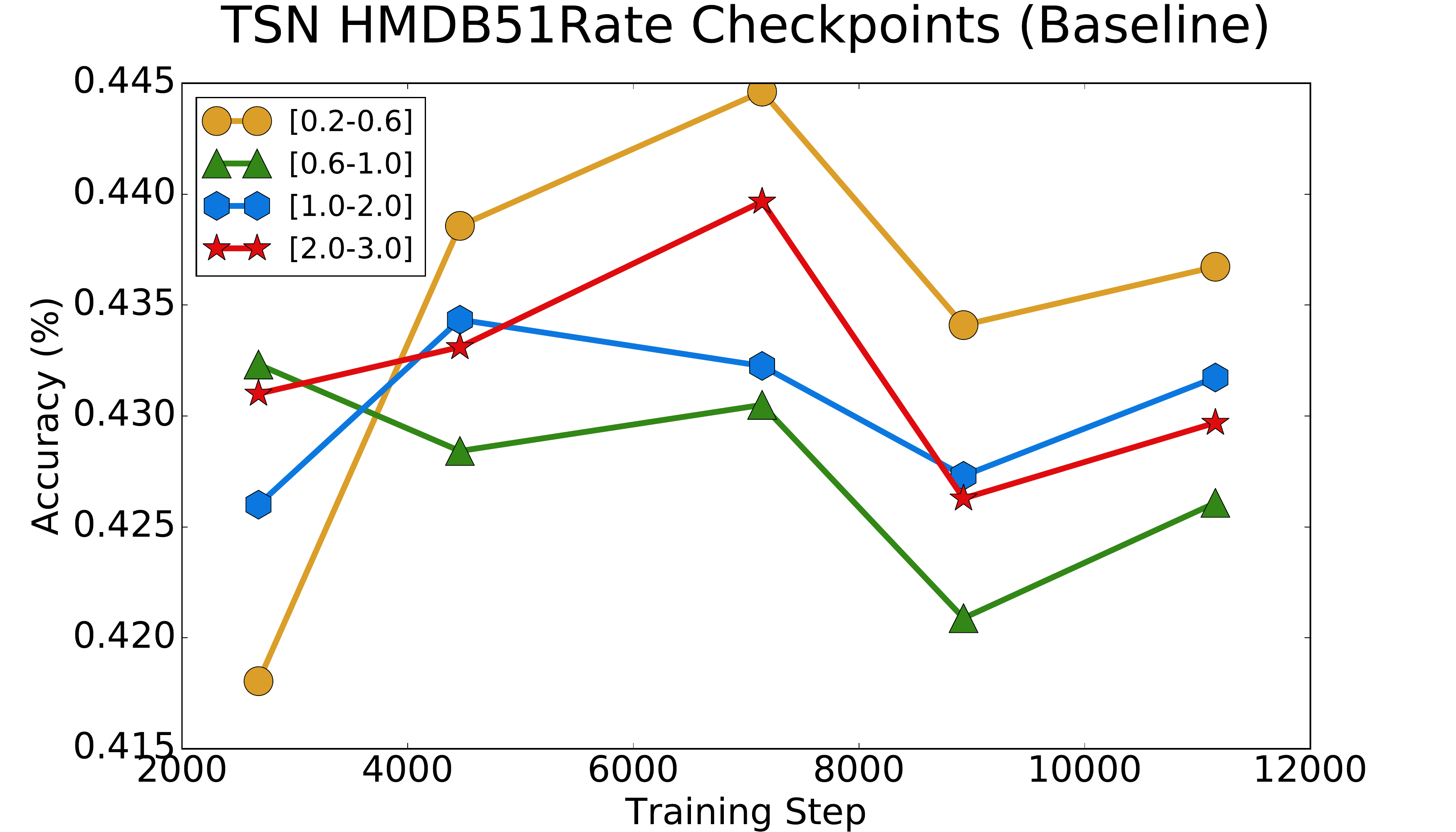}
  \end{subfigure}
  \vspace{1cm}
  \begin{subfigure}[h]{0.49\textwidth}
    \centering
    \includegraphics[width=\textwidth]{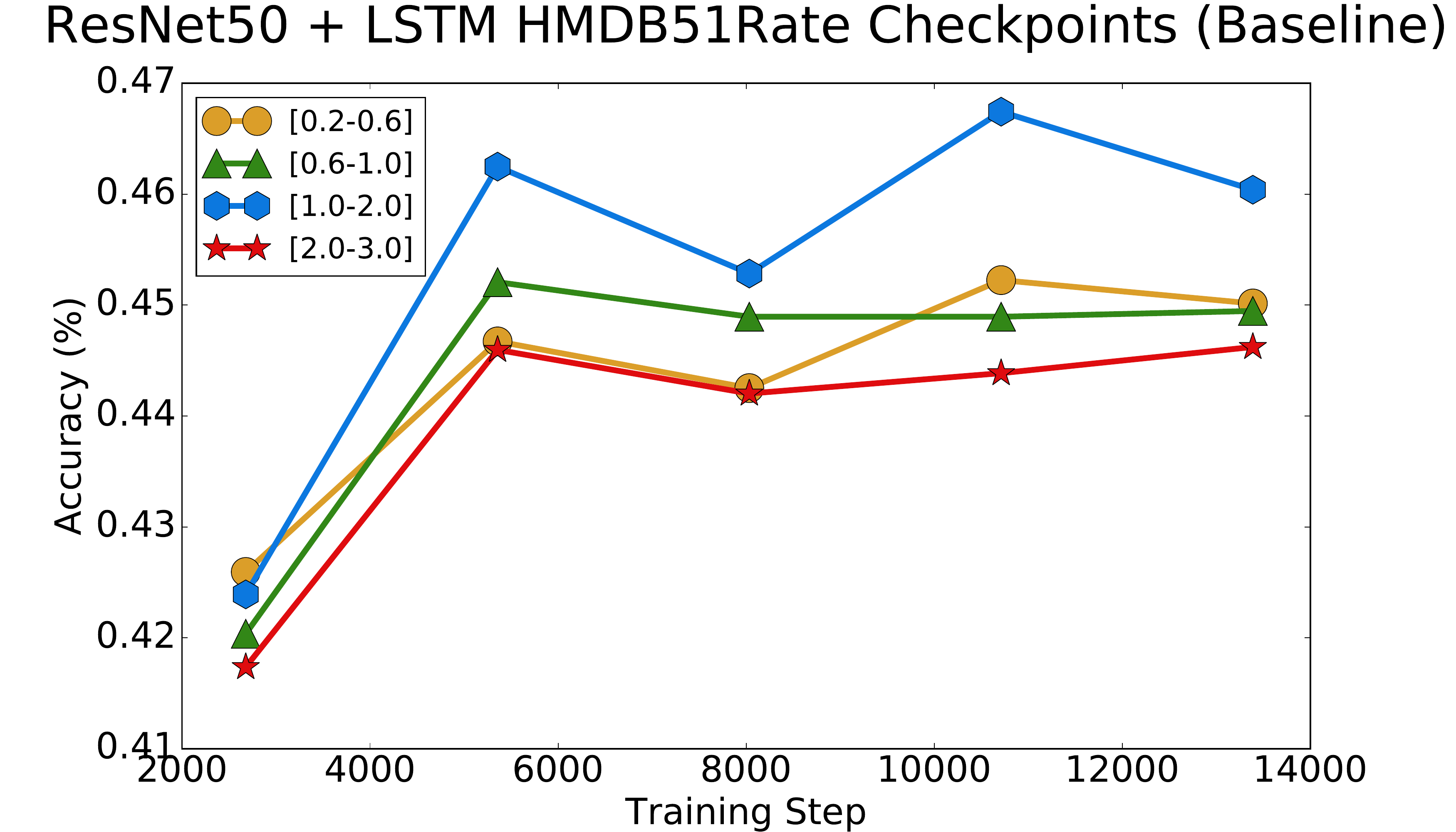}
  \end{subfigure}
  \vspace{1cm}
  \begin{subfigure}[h]{0.49\textwidth}
    \centering
    \includegraphics[width=\textwidth]{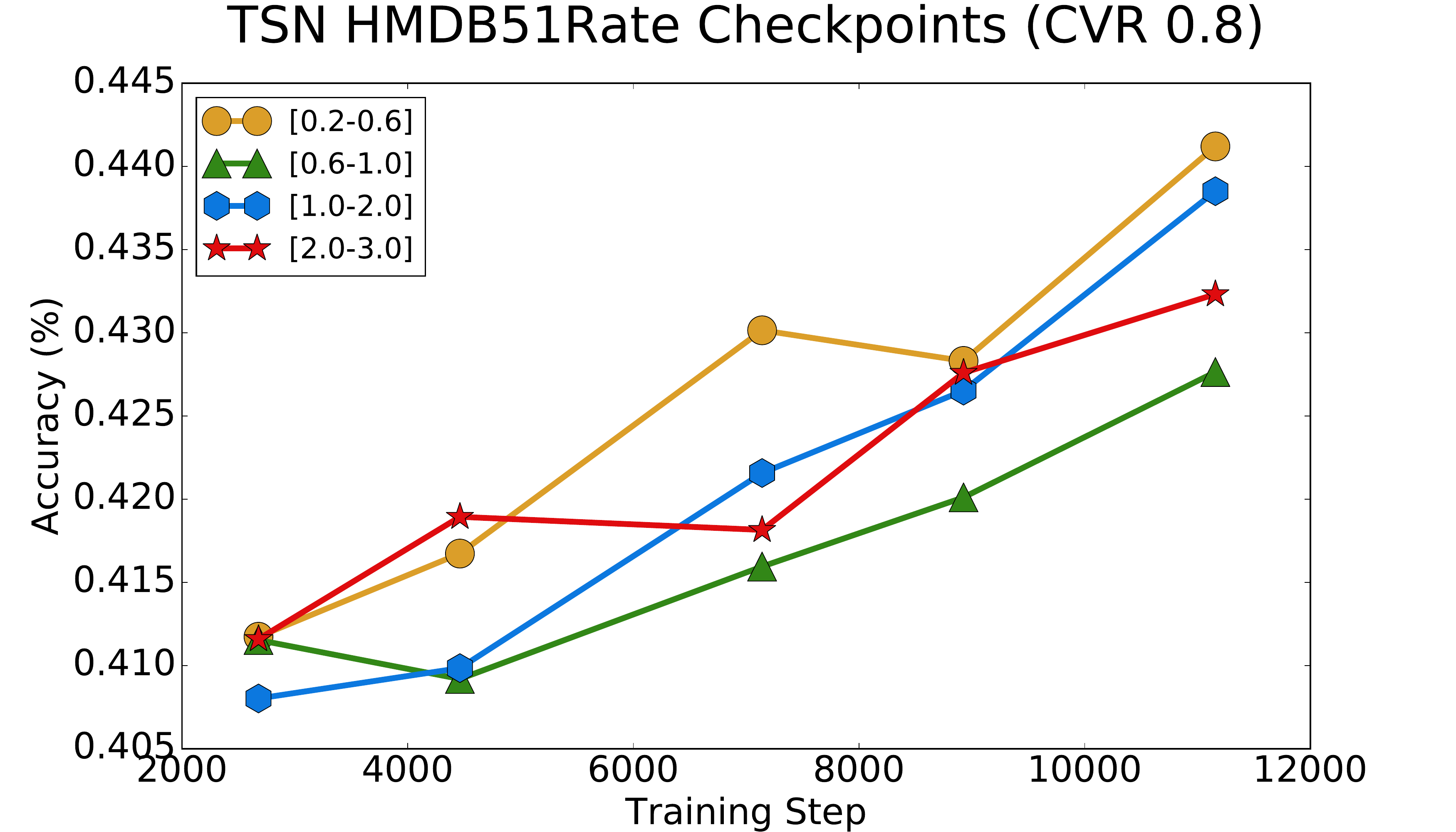}
  \end{subfigure}
  \begin{subfigure}[h]{0.49\textwidth}
    \centering
    \includegraphics[width=\textwidth]{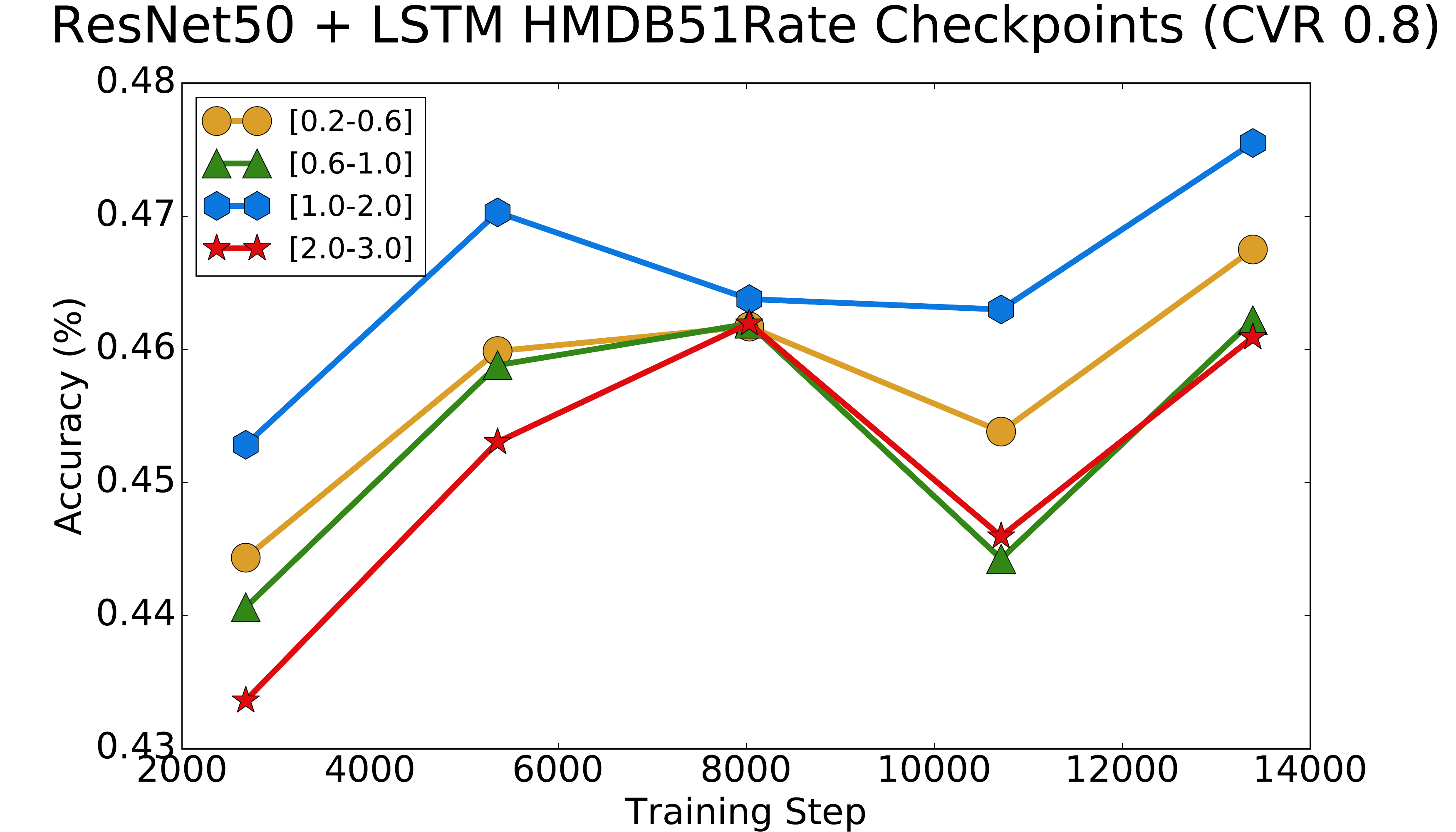}
  \end{subfigure}
  \begin{subfigure}[h]{0.49\textwidth}
    \centering
    \includegraphics[width=\textwidth]{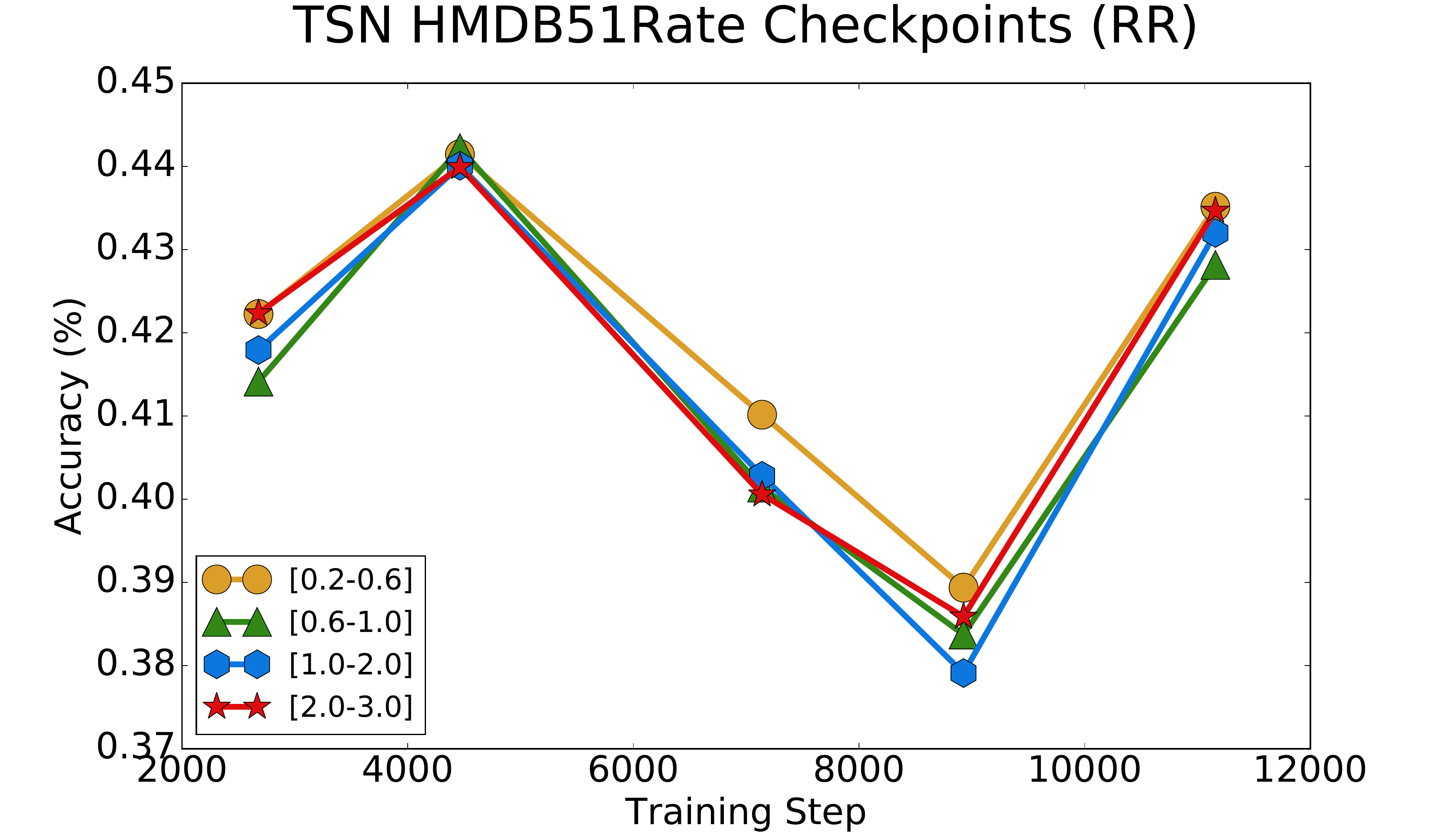}
  \end{subfigure}
  \begin{subfigure}[h]{0.49\textwidth}
    \centering
    \includegraphics[width=\textwidth]{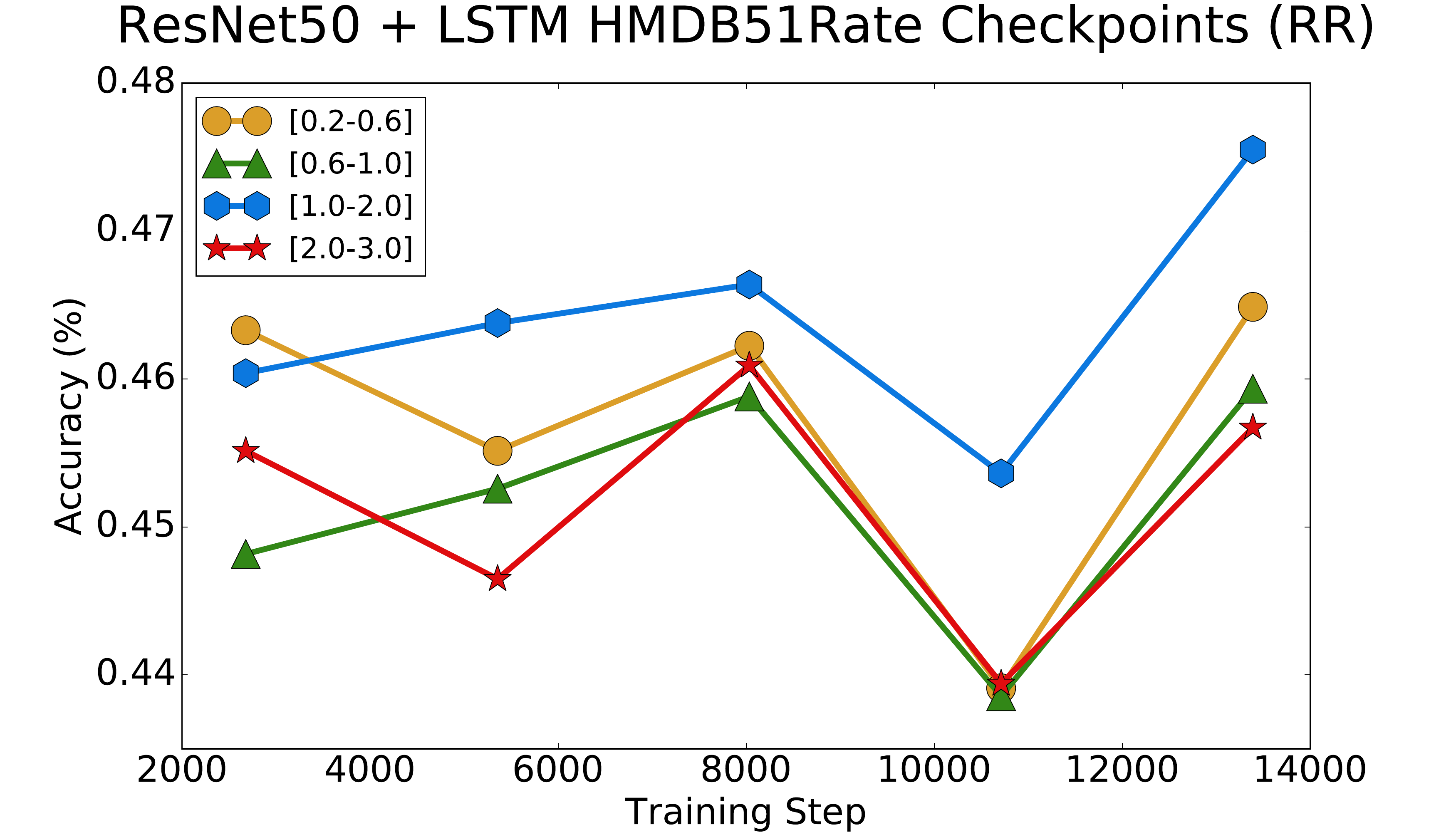}
  \end{subfigure}
  \caption{TSN and ConvNet + LSTM performance across the HMDB51Rate dataset is shown for five training checkpoints. 
The HMDB51Rate dataset is split into four bins of resampling factors: [$0.2-0.6$, $0.6-1.0$, $1.0-2.0$, $2.0-3.0$]. 
For both TSN and ConvNet + LSTM, we test the baseline and a static and dynamic \acro~model. 
The selected \acro~models are those with the highest performance from our input $\alpha$ testing.}
  \label{fig:type_2_checkpoints}
\end{figure}

The temporal characteristics of Type II models are less well defined than those of Type I models. 
C3D and I3D show similar behaviors to our input $\alpha$ testing across their \acro-based variants while TSN and ConvNet + LSTM \acro-based models show different reactions to input $\alpha$ tests between one another, as seen in Figs.~\ref{fig:tsn_resnet_complete_alpha_testing}.
In order to determine the evolution of robustness to videos of varying speed in our Type II models, we test them on the HMDB51Rate dataset at five different epochs throughout their training phase.
This dataset modifies the original HMDB51 dataset in that every video is resampled using ten different randomly selected resampling factors.
The resampling factors are evenly sampled across the four bins, [$0.2-0.6, 0.6-1.0, 1.0-2.0, \text{ and } 2.0-3.0$].
For both Type II models, we test the baseline, CVR-$0.8$, and RR variants.
   
Fig.~\ref{fig:type_2_checkpoints} shows the results of this checkpoint testing for TSN models in the left-hand column and ConvNet + LSTM models in the right-hand column.
The TSN model variants show extremely different learning behavior when compared to each other.
We see that the baseline model shows least stability between the performances across various resampling factor bins.
On the other hand, TSN CVR-0.8 shows an almost steady improvement in performance across all bins throughout its training phase.
An important characteristic directly observable from these plots is that the \acro-based TSN models show very similar behaviour in terms of recognition performance across resampling factor bins within each model. 
This is not the case for the baseline TSN model.

When compared to all of our ConvNet + LSTM models, the baseline and \acro~variants, perform similarly to one another in terms of performance fluctuations between rate bins during training.
We hypothesise the reason that the ConvNet + LSTM models show a lesser impact when applying \acro~preprocessing is due to the inherent temporal modeling performed within the LSTM, which gives such models their high stability characteristic.

\end{document}